\PassOptionsToPackage{table,dvipsnames}{xcolor}
\documentclass[10pt,twocolumn,letterpaper]{article}
\usepackage{amsmath}
\usepackage{tabularx} 

\usepackage{listings} 
\usepackage{makecell}

\usepackage{multirow} %
\usepackage{booktabs}  %
\usepackage{graphicx}  %
\usepackage{xcolor}
\usepackage[accsupp]{axessibility}
\usepackage{accessibility}

\usepackage[T1]{fontenc}
\usepackage{times}
\usepackage{iccv}              

\definecolor{iccvblue}{rgb}{0.21,0.49,0.74}
\usepackage[pagebackref,breaklinks,colorlinks,allcolors=iccvblue]{hyperref}

\def\paperID{9224} 
\def\confName{ICCV}
\def\confYear{2025}

\title{GCAV: A Global Concept Activation Vector Framework for Cross-Layer Consistency in Interpretability}

\author{
Zhenghao He, Sanchit Sinha, Guangzhi Xiong, Aidong Zhang \\
University of Virginia, USA \\
{\tt\small \{zhenghao, sanchit, guangzhi, aidong\}@virginia.edu}
}

\begin{document}
\maketitle

\begin{abstract}
Concept Activation Vectors (CAVs) provide a powerful approach for interpreting deep neural networks by quantifying their sensitivity to human-defined concepts. However, when computed independently at different layers, CAVs often exhibit inconsistencies, making cross-layer comparisons unreliable. To address this issue, we propose the Global Concept Activation Vector (GCAV), a novel framework that unifies CAVs into a single, semantically consistent representation. Our method leverages contrastive learning to align concept representations across layers and employs an attention-based fusion mechanism to construct a globally integrated CAV. By doing so, our method significantly reduces the variance in TCAV scores while preserving concept relevance, ensuring more stable and reliable concept attributions. To evaluate the effectiveness of GCAV, we introduce Testing with Global Concept Activation Vectors (TGCAV) as a method to apply TCAV to GCAV-based representations. We conduct extensive experiments on multiple deep neural networks, demonstrating that our method effectively mitigates concept inconsistency across layers, enhances concept localization, and improves robustness against adversarial perturbations. By integrating cross-layer information into a coherent framework, our method offers a more comprehensive and interpretable understanding of how deep learning models encode human-defined concepts. Code and models are available at 
\url{https://github.com/Zhenghao-He/GCAV}.
\end{abstract}
    
\section{Introduction}
\label{sec:intro}

Deep learning models have achieved remarkable success in various domains, yet their ``black-box'' nature raises challenges for trust, accountability, and deployment in critical applications. To address this, interpretability methods aim to provide insight into model decision making. Traditionally, researchers have analyzed feature importance by examining input attributions. However, such methods struggle to explain how models develop a higher-level conceptual understanding beyond raw features.

Concept Activation Vectors (CAVs) offer a powerful alternative by enabling interpretability in terms of human-understandable concepts~\cite{kim2018interpretability}. By training a linear classifier to distinguish between instances that contain a specific concept and those that do not, CAVs define semantic directions in the model’s latent space, allowing researchers to quantify concept influence using Testing with Concept Activation Vectors (TCAV).


Although TCAV provides valuable information, it faces a major limitation: \textbf{CAVs are computed independently for each layer}, leading to inconsistent interpretations across layers. This introduces several challenges:
\begin{itemize}
\item \textbf{Unstable layer selection}: The same concept may have varying importance across layers, making it unclear which layer should be used for interpretation.
\item \textbf{Spurious activations}: Certain layers assign high TCAV scores to irrelevant concepts (e.g., \emph{dotted, zagzagged} in class \emph{zebra}), reducing reliability.
\item \textbf{High variance in TCAV scores}: Due to inconsistencies in concept representation across layers, TCAV scores fluctuate significantly, making the results difficult to trust.
\end{itemize}


These challenges arise because deep neural networks process information hierarchically~\cite{zeiler2014visualizing,krizhevsky2009learning,yamins2014performance,bengio2013representation}—lower layers tend to capture low-level features (e.g., edges and textures), whereas deeper layers encode high-level semantic concepts. Consequently, computing CAVs independently at each layer ignores the fact that different layers contribute differently to the concept representation. Furthermore, model architectures vary significantly in depth and feature extraction mechanisms. For example, ResNet~\cite{he2016identity} has substantially more layers than MobileNet~\cite{sandler2018mobilenetv2}, meaning that a layer selection strategy tailored to one architecture may not generalize well to another.


To address these issues, we propose \textbf{Global Concept Activation Vector (GCAV)}, a novel framework that unifies concept representations across layers into a globally consistent CAV. Specifically, GCAV uses contrastive learning~\cite{oord2018representation} to align cross-layer representations and an attention-based fusion mechanism~\cite{vaswani2017attention} to aggregate information. By integrating information from multiple layers, GCAV constructs a globally integrated concept representation that remains stable across layers, reducing variance in TCAV scores, and ensuring more robust and semantically consistent model interpretation.

Our main contributions are summarized as follows:

\begin{itemize}
\item \textbf{Globally Consistent Concept Representation}: We propose a novel framework that integrates CAVs from multiple layers into a single unified representation. This ensures that the importance of the concept remains consistent across layers, eliminating ambiguity in layer selection, and reducing the variance of TCAV scores.

\item \textbf{Reduction of Spurious Concept Activations}: Our method suppresses the undesired activation of irrelevant concepts in TCAV scores. Unlike the original TCAV method, where certain layers exhibit high scores even for concepts not related to the target class, our framework ensures that only semantically meaningful concepts receive high attributions. 

\item \textbf{Enhanced Interpretability and Robustness}: By integrating concept representations across layers, GCAV reduces sensitivity to adversarial attacks. Unlike TCAV, which is vulnerable to perturbations in specific layers, the globally consistent representation of GCAV stabilizes the importance of the concept, making interpretability more robust~\cite{Brown2021MakingCI}.
\end{itemize}

\section{Related Work}
\label{sec:related}
Explaining the decision-making process of deep neural networks (DNNs) is a challenging and active area of research. Existing methods can be broadly classified into three major paradigms: feature-based, sample-based, and concept-based. Feature-based methods assign importance scores to individual input features, as seen in approaches such as LIME~\cite{ribeiro2016should}, Integrated Gradients~\cite{sundararajan2017axiomatic}, GradCAM~\cite{selvaraju2017grad}, and SHAP~\cite{lundberg2017unified}. Sample-based methods instead focus on identifying influential training samples for a given prediction, using techniques like Influence Functions~\cite{koh2017understanding,feldman2020neural} and Hessian-based approximations~\cite{pruthi2020estimating,yeh2018representer}.  
 
Concept-based methods aim to provide human-aligned explanations by associating model predictions with high-level semantic concepts. Unlike feature attribution methods that assign importance scores to individual pixels or neurons~\cite{ribeiro2016should, sundararajan2017axiomatic, selvaraju2017grad, lundberg2017unified}, concept-based approaches focus on understanding how abstract concepts influence model behavior.  

Several techniques have been proposed to incorporate concepts into model interpretability. Some methods attempt to discover meaningful concepts automatically, such as Automatic Concept-based Explanations (ACE)~\cite{ghorbani2019towards}, which clusters input features into semantically distinct concepts. Others integrate concept supervision during training, as seen in Concept Bottleneck Models (CBMs)~\cite{koh2020concept}, where an intermediate layer is explicitly trained to represent human-defined concepts. While these approaches enhance interpretability by introducing structured concept representations, they do not provide a direct way to quantify the influence of a given concept on model predictions.  

To bridge this gap, Testing with Concept Activation Vectors (TCAV)~\cite{kim2018interpretability} was introduced as a post-hoc method for assessing a model’s sensitivity to predefined concepts. TCAV quantifies concept importance by computing directional derivatives along concept activation vectors (CAVs) in the model’s latent space, providing a global measure of concept influence. However, despite its effectiveness, TCAV has several limitations. First, TCAV explanations are highly sensitive to the choice of model layers, leading to inconsistencies across layers~\cite{nicolson2024explaining}. Second, TCAV is vulnerable to adversarial perturbations, where small modifications to activations at a single layer can significantly alter concept importance scores~\cite{Brown2021MakingCI}.  

To address these issues, prior research has explored improving TCAV’s reliability by reducing concept correlation~\cite{goyal2019explaining, chen2020concept}, refining concept attributions through data augmentation~\cite{anders2022finding}, or enforcing structured concept learning~\cite{gupta2023concept}. While these methods enhance concept representation, they do not resolve TCAV’s fundamental issue of layer-wise inconsistency, which remains a major limitation in post-hoc concept-based explanations.

To overcome these limitations, we propose Global Concept Activation Vector (GCAV), a novel framework that integrates CAVs across multiple layers into a semantically consistent representation. Unlike TCAV, which computes layer-specific concept directions independently, GCAV constructs a unified representation that improves cross-layer consistency and robustness.

\section{Methodology}
In this section, we introduce our method: (a) how to get a unified Global Concept Activation Vector (GCAV), which is not layer sensitive, and (b) how to Test with Global Concept Activation Vector (TGCAV).
\subsection{Overview of the Integration Framework}
\begin{figure*}[h]
  \centering %
  \includegraphics[width=\linewidth]{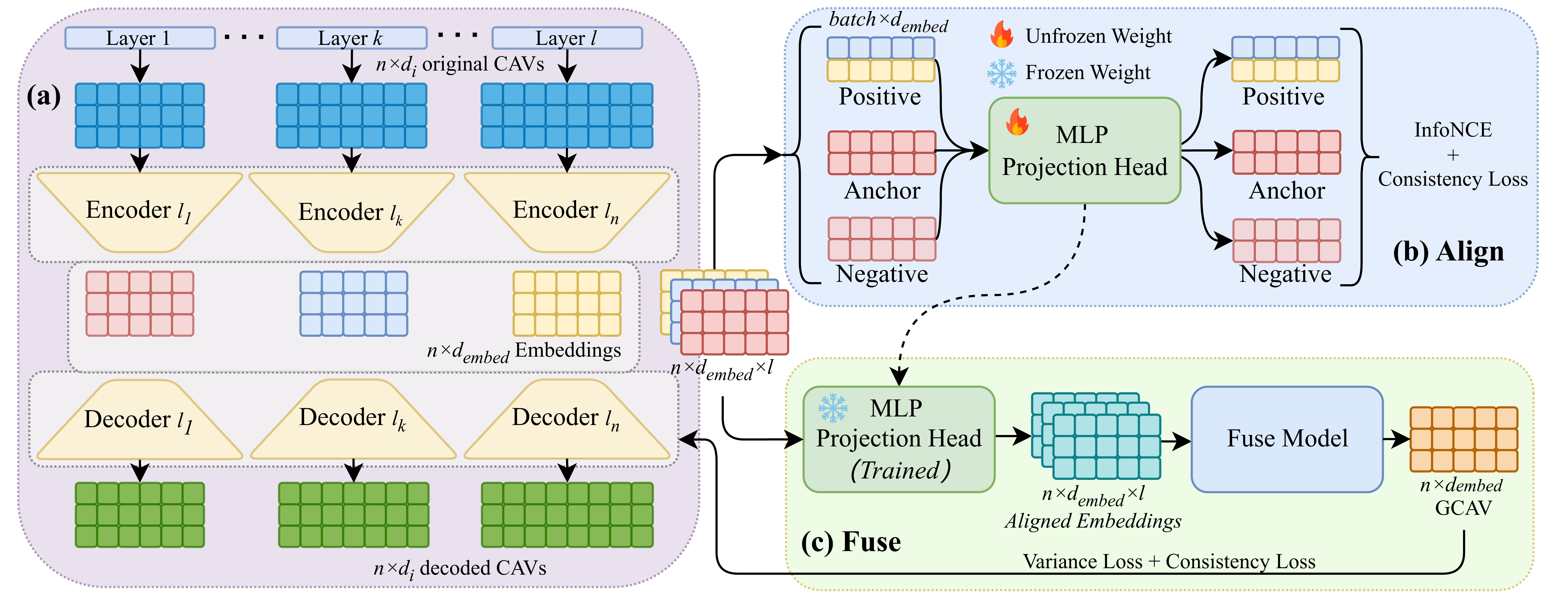}
  \caption{\textbf{Framework of training GCAV}: The proposed framework consists of three training stages (a)–(c), which are trained sequentially. (a) \textbf{Layer-wise Autoencoder Training}: Each layer’s high-dimensional CAVs are compressed into a unified embedding space via independently trained autoencoders. (b) \textbf{Cross-Layer Alignment}: Contrastive learning is applied using an MLP projection head to align embeddings across layers, leveraging InfoNCE and consistency loss. (c) \textbf{Global CAV Fusion}: The aligned embeddings are fused into GCAV using an attention-based fusion model, optimized with variance and consistency loss.}\label{framework}
\end{figure*}

In our work, we propose an integration framework designed to address the layer-wise inconsistency of Concept Activation Vectors (CAVs) and to achieve global semantic consistency in model interpretability. As can be seen in Figure \ref{framework}, our framework consists of three key components: (1) layer-specific autoencoders for dimensional unification, (2) contrastive learning for aligning CAVs across layers, and (3) an attention-based fusion mechanism for synthesizing a GCAV. 

Our framework consists of three sequential training stages, while one may consider jointly training these components, we adopt a stage-wise optimization strategy due to the following reasons:

\noindent \textbf{Autoencoder vs. Align\&Fuse (Stage 1 vs. Stages 2/3).}  
Autoencoders aim to reduce CAV dimensionality while preserving concept-specific information, whereas contrastive learning enforces alignment across layers. If trained together, contrastive loss would interfere with the autoencoder’s compression objective, forcing it to prioritize inter-layer similarity over preserving information. This could lead to early collapse, where all embeddings become too similar across layers, reducing interpretability. 

Similarly, the decoder is designed to reconstruct the original CAV embeddings, ensuring that per-layer information is retained. However, if trained jointly with the fusion model, the decoder may adapt to the fusion objective rather than faithfully reconstructing each layer’s CAV. This would result in fusion learning from biased embeddings that do not accurately represent the original layer-wise information, compromising interpretability.  

\noindent \textbf{Align vs. Fuse (Stage 2 vs. Stage 3).}  
Contrastive learning focuses on aligning layer-specific embeddings, but it does not directly construct a unified global CAV. If fusion is trained simultaneously, it might learn from embeddings that are not yet well-aligned, leading to an inconsistent global representation. Delaying fusion until after alignment stabilizes ensures that the final representation is semantically coherent across layers.

By following a progressive optimization approach—first compressing high-dimensional CAVs, then aligning them, and finally synthesizing a unified CAV—we ensure a robust, interpretable, and semantically consistent representation.

\subsection{Dimension Unification}
\label{sec:dimension unification}
Each layer of the network produces high-dimensional CAVs that vary in both dimension and semantic representation, which poses challenges for subsequent training. Additionally, the original CAV dimension is determined by the activation values of the model being explained, and these dimensions can be very large, making direct computation time-consuming. To address these issues, we train layer-specific autoencoders to standardize the dimensionality of CAVs while keeping them in independent latent spaces.
For each layer \( l \), we train a dedicated autoencoder to project the high-dimensional CAV \( x_{l}^c \) of concept \(c\) into a latent space of fixed dimension. The encoder for layer \( l \), denoted as \( f_{\text{encoder}}^l \), computes the embedding as:
\begin{equation}
\label{eq:encoder}
 z_{l}^c = f_{\text{encoder}}^l(x_{l}^c),
 \quad x_{l}^c \in \mathbb{R}^{d_c}, 
 \quad z_l^c \in \mathbb{R}^{d_{\text{embed}}}
\end{equation}
Although each layer has its own autoencoder, the embeddings \( \mathbf{z}_{l}^c \) are of the same dimension \( d_{\text{embed}} \), ensuring a unified embedding across layers. 
The corresponding decoders are computed as follows \( f_{\text{decoder}}^l \):
\begin{equation}
\label{eq:decoder}
\tilde{x}_{l}^c = f_{\text{decoder}}^l(z_{l}^c), 
\quad \tilde{x}_{l}^c \in \mathbb{R}^{d_c}
\end{equation}
These autoencoders are trained by minimizing the reconstruction error, which is measured as the cosine similarity between the reconstructed CAV \(\tilde{x}_{l}^c\) and the original CAV \(x_{l}^c\):
\begin{equation}
    \mathcal{L}_{\text{rec}} = 1 - \frac{\tilde{x}_{l}^c \cdot x_{l}^c}{\|\tilde{x}_{l}^c\| \|x_{l}^c\|}
\end{equation}
where a lower \(\mathcal{L}_{\text{rec}}\) indicates better reconstruction quality.

Constructing this autoencoder has the following advantages:
\begin{itemize}
\item {Unified Dimensionality}: Ensures that CAVs from various layers are aligned in a consistent embedding space, simplifying subsequent training.
\item {Dimensionality Reduction}: Reducing the dimensions decreases computation time and resource requirements, while preserving essential features.
\item {Reconstruction for TGCAV Evaluation}: Uses trained decoders to project low-dimensional representations back to each layer, enabling TGCAV computation and evaluation of GCAV.
\end{itemize}


\subsection{Contrastive Learning for CAV Alignment}
\label{sec:Contrastive Learning for CAV Alignment}
We denote the embedding of concept \( c \) at layer \( l \), generated by its autoencoder, as \( \mathbf{z}_{l}^{c} \). Since these embeddings are obtained from different autoencoders at different layers and may possess distinct semantic characteristics, it is necessary to align them into a shared space. To achieve this, we employ contrastive learning to refine the representations of the same predefined concept across different layers.  

To facilitate this alignment, we introduce a projection function \( f(\cdot) \), implemented as a multi-layer perceptron (MLP) with residual connections. It transforms CAV embeddings \(z_l^c\) into a unified space, ensuring that representations from different layers are comparable while preserving essential semantic properties (see Appendix for details).

As illustrated in Figure \ref{framework}, our MLP projection head takes three types of inputs: \textit{Anchor} (\(\mathbf{z}_a\)), \textit{Positive} (\(\mathbf{z}_p\)), and \textit{Negative} (\(\mathbf{z}_n\)). In our approach, \textit{positive pairs} consist of CAV embeddings corresponding to the same concept \( c \) but extracted from different layers \( l_1, l_2 \), i.e.,  
\begin{equation}
(\mathbf{z}_{a}, \mathbf{z}_{p}) \in \{ (\mathbf{z}_{l_1}^{c}, \mathbf{z}_{l_2}^{c}) \mid l_1 \neq l_2 \}.
\end{equation}
Conversely, negative samples are obtained by computing CAVs from two entirely random datasets. The corresponding \textit{negative pairs} consist of CAV embeddings extracted from the same layer \(l\), but the negative CAVs are trained using two random probe datasets. That is,
\begin{equation}
(\mathbf{z}_{a}, \mathbf{z}_{n}) \in \{ (\mathbf{z}_{l}^{c}, \mathbf{z}_{l}^{r_1, r_2}) \mid l \in \mathcal{L} \},
\end{equation}
where \( r_1 \) and \( r_2 \) denote two distinct random datasets, ensuring that the CAVs are derived from data that does not contain any concept-specific information.

The \textbf{InfoNCE loss} \cite{oord2018representation} is used to align CAV embeddings across layers. For an anchor \(\mathbf{z}_a\) and its positive pair \(\mathbf{z}_p\) (extracted from different layers for the same concept), and a set of negatives \(\{\mathbf{z}_{n}\}\) (derived from random datasets), it is defined as:
\begin{equation} \small
\begin{aligned}
\mathcal{L}_{\mathrm{NCE}} = & -\log \frac{\exp\big(\operatorname{sim}(\mathbf{z}_a, \mathbf{z}_p)/\tau\big)}
{\exp\big(\operatorname{sim}(\mathbf{z}_a, \mathbf{z}_p)/\tau\big) + \sum\limits_{\mathbf{z}_n} \exp\big(\operatorname{sim}(\mathbf{z}_a, \mathbf{z}_n)/\tau\big)},
\end{aligned}
\end{equation}
where \(\operatorname{sim}(\cdot,\cdot)\) denotes a similarity function (in this paper, we applied cosine similarity) and \(\tau\) is the temperature parameter.

The \textbf{Consistency loss} ensures that the transformation performed by \( f(\cdot) \) does not distort concept-specific information. Specifically, given an original CAV embedding \(\mathbf{z}^{c}\), we define:
\begin{equation}
\label{eq:consistency loss}
    \mathcal{L}_{\mathrm{cons}} = 1 - \frac{\mathbf{z}^{c} \cdot f(\mathbf{z}^{c})}{\|\mathbf{z}^{c}\| \|f(\mathbf{z}^{c})\|}
\end{equation}
where \( f(\cdot) \) is the projection function. This loss encourages \( f(\cdot) \) to retain the critical semantic details while mapping CAV embeddings into a consistent space. 
In our training process, we jointly minimize an overall loss:
\begin{equation}
\mathcal{L} = \lambda_{\mathrm{NCE}}\,\mathcal{L}_{\mathrm{NCE}} + \lambda_{\mathrm{cons}}\, \mathcal{L}_{\mathrm{cons}},
\end{equation}
where \(\lambda_{\mathrm{NCE}}\) and \(\lambda_{\mathrm{cons}}\)are hyperparameters that balance the two terms.
Together, these losses ensure that embeddings from different layers for the same concept are merged into a unified semantic space while preserving their interpretability.

\subsection{Cross-Layer CAV Fusion}
\label{sec:Cross-Layer CAV Fusion}
To aggregate information from CAVs obtained at different layers, we adopt a fusion module based on Transformer-style self-attention. Let 
\begin{equation}
\mathbf{Z} = \{\tilde{\mathbf{z}^c_l} \in \mathbb{R}^{d_{\text{embed}}} \mid l=1,\ldots,L\}
\end{equation}
be the set of aligned CAV embeddings \(\tilde{\mathbf{z}^c_l}\) extracted from \(L\) different layers for concept \(c\). we introduce a learnable positional encoding to incorporate the layer index, allowing the model to capture layer-wise dependencies. These embeddings are then processed using self-attention to dynamically reweight information from different layers, ensuring that the final representation captures a globally consistent concept representation. Finally, an average pooling over the layer dimension yields the fused GCAV:
\begin{equation}
\mathbf{z}_{\text{GCAV}}^c = \frac{1}{L} \sum_{l=1}^{L} \tilde{\mathbf{z}^c_l},
\quad \mathbf{z}_{\text{GCAV}}^c \in \mathbb{R}^{d_{\text{embed}}}
\end{equation}
where \(\mathbf{z}_{\text{GCAV}}^c\) is the output of our Transformer-based fusion module that aggregates the aligned CAV embeddings for concept \(c\) from all layers. This GCAV retains the same dimensionality \(d_{\text{embed}}\) as defined in Equation~\ref{eq:encoder}, and captures a unified semantic representation of concept \(c\) independent of any specific layer.

\noindent\textbf{Loss Function Design}.
Let \(\mathbf{z}_{\text{GCAV}}^c \in \mathbb{R}^{d_{\text{embed}}}\) be the GCAV for a given concept \(c\), and let \(f_{\text{decoder}}^l\) denote the decoder corresponding to layer \(l\). The reconstructed CAV at layer \(l\) is given by:
\begin{equation}
    \tilde{\mathbf{v}_l^c} = f_{\text{decoder}}^l\bigl( \mathbf{z}_{\text{GCAV}}^c \bigr).
\label{eq:project back}
\end{equation}
Then We replace the original CAV with \(\tilde{\mathbf{v}_l^c}\) and compute the TCAV score \(s_l\) on layer \(l\).

We define the Variance Loss \(\mathcal{L}_{\text{var}}\) to minimize the variance of TCAV scores across layers over a batch of \(N\) samples:
\begin{equation}
    \mathcal{L}_{\text{var}} = \frac{1}{N} \sum_{i=1}^{N} \operatorname{Var}\Bigl( s^{(i)}_1, s^{(i)}_2, \dots, s^{(i)}_L \Bigr).
    \label{eq:variance loss}
\end{equation}

To ensure that the fusion function \(g(\cdot)\) preserves the semantic content of the original CAVs, we also introduce the Consistency Loss:
\begin{equation}
    \mathcal{L}_{\mathrm{cons}} = 1 - \frac{\tilde{\mathbf{v}_l^c} \cdot f_{\text{decoder}}^l\bigl( \tilde{\mathbf{z}^c_l}\bigr)}{\|\tilde{\mathbf{v}_l^c}\| \|f_{\text{decoder}}^l\bigl( \tilde{\mathbf{z}^c_l}\bigr)\|}
\end{equation}
where \(\tilde{\mathbf{v}_l^c} \in \mathbb{R}^{d_{\text{embed}}}\) is the GCAV projected back to layer \(l\), representing the concept \(c\). \(\tilde{\mathbf{z}^c_l} \in \mathbb{R}^{d_{\text{embed}}}\) is the aligned CAV embedding at layer \(l\). And \(f_{\text{decoder}}^l(\cdot)\) is the decoder function 

By minimizing this loss, we encourage the fusion model to retain concept-relevant information across layers. This ensures that, after projection, the reconstructed CAVs remain semantically meaningful and close to their original representations.

The final loss function is defined as a weighted sum of the variance and consistency losses:
\begin{equation}
    \mathcal{L} = \lambda_{\text{var}}\, \mathcal{L}_{\text{var}} + \lambda_{\text{cons}}\, \mathcal{L}_{\text{cons}},
\end{equation}
where \(\lambda_{\text{var}}\) and \(\lambda_{\text{cons}}\) are hyperparameters that balance the two terms.

\subsection{Testing with GCAV (TGCAV)}

To evaluate the impact of a concept at different layers, we introduce Testing with Global Concept Activation Vectors (TGCAV). Specifically, we first obtain a global concept representation \( \mathbf{z}_{\text{GCAV}}^c \). This representation is then projected back into each layer’s feature space using the corresponding trained decoder \( f_{\text{decoder}}^l \) as described in Section~\ref{sec:dimension unification}, yielding a layer-specific concept vector that remains semantically aligned with the global representation. By applying the standard TCAV procedure to these reconstructed representations, we obtain TGCAV scores that reflect the influence of a concept at different layers while preserving semantic consistency.

Formally, the TGCAV score at layer \( l \) for concept \( c \) and class \( k \) is defined as:
\begin{equation} \footnotesize
    TGCAV_{c, k, l} = \frac{\big| \{ x \in X_k : \nabla h_{l,k} ( f_l(x) ) \cdot f_{\text{decoder}}^l ( \mathbf{z}_{\text{GCAV}}^c )  > 0 \} \big|}{|X_k|},
\end{equation}
where \(X_k\) is the set of inputs belonging to class \(k\), \(f_l(x)\) denotes the activation at layer \(l\) for input \(x\), and \(h_{l,k} : \mathbb{R}^m \to \mathbb{R}\) maps these activations to the logit for class \(k\).

\section{Experiments}
\label{sec:Experiments}
In this section, we demonstrate the result of applying our method to state-of-art convolutional networks. We show that our method \( (\text{i}) \) eliminates layer-wise variability and achieves global semantic consistency, \( (\text{ii}) \) improves the precision of concept localization, and \( (\text{iii}) \) provides a more robust explanation approach that is less susceptible to adversarial attacks.

\subsection{Experimental Setup}
Our experiments are conducted on ResNet50V2 \cite{he2016identity}, GoogleNet \cite{szegedy2015going}, and MobileNetV2 \cite{sandler2018mobilenetv2}, all of which are pre-trained on the ImageNet \cite{5206848} dataset. For each concept, we select 50 images from the Broden dataset \cite{netdissect2017}, consistent with the original TCAV paper. We perform 10 random experiments for each concept, using 10 different random probe datasets, each containing 50 random images from ImageNet, to train the linear model and compute the Concept Activation Vectors (CAVs). 

We train our linear models to get original CAVs with the default hyperparameters~\cite{kim2018interpretability} in the original TCAV paper's code. The alignment phase optimizes both InfoNCE loss and consistency loss, with a weighting ratio of \(\lambda_{\mathrm{NCE}} : \lambda_{\mathrm{cons}} = 1:3\). In the fusion phase, we balance variance loss and consistency loss to ensure robust global CAV representation, using a weighting ratio of \(\lambda_{\mathrm{var}} : \lambda_{\mathrm{cons}} = 3:1\).

\subsection{Layer-wise Distribution Analysis}

\begin{table*}[ht]
\centering
\begin{tabularx}{\textwidth}{p{0.8cm} p{1.4cm} p{1.0cm} *{3}{*{4}{X}}}
\toprule
\multirow{2.5}{*}{Target} & \multirow{2.5}{*}{Concept} & \multirow{2.5}{*}{Method} 
& \multicolumn{4}{c}{GoogleNet} 
& \multicolumn{4}{c}{ResNet50V2} 
& \multicolumn{4}{c}{MobileNetV2} \\
\cmidrule(lr){4-7} \cmidrule(lr){8-11} \cmidrule(lr){12-15}
& & 
& Mean & Std & CV & RR  
& Mean & Std & CV & RR  
& Mean & Std & CV & RR  \\
\midrule
\multirow{6}{*}{\makecell{Spider\\Web}} 
 & \multirow{2}{*}{cobwebbed} & TCAV  
 & 0.808 & 0.140 & 0.173 & 0.508  
 & 0.644 & 0.127 & 0.197 & 0.715
 & 0.619 & 0.169 & 0.273 & 0.985\\
 &  & \cellcolor{blue!15} TGCAV  
 & \cellcolor{blue!15} 0.840 & \cellcolor{blue!15} 0.109 & \cellcolor{blue!15} 0.130 & \cellcolor{blue!15} 0.381  
 & \cellcolor{blue!15} 0.784 & \cellcolor{blue!15} 0.083 & \cellcolor{blue!15} 0.105 & \cellcolor{blue!15} 0.357 
 & \cellcolor{blue!15} 0.554 & \cellcolor{blue!15} 0.091 & \cellcolor{blue!15} 0.164 & \cellcolor{blue!15} 0.686\\ 
 & \multirow{2}{*}{porous} & TCAV  
 & 0.499 & 0.123 & 0.246 & 0.862  
 & 0.586 & 0.098 & 0.167 & 0.717
 & 0.377 & 0.059 & 0.157 & 0.584\\
 &  & \cellcolor{blue!15} TGCAV  
 & \cellcolor{blue!15} 0.498 & \cellcolor{blue!15} 0.037 & \cellcolor{blue!15} 0.074 & \cellcolor{blue!15} 0.281  
 & \cellcolor{blue!15} 0.640 & \cellcolor{blue!15} 0.030 & \cellcolor{blue!15} 0.047 & \cellcolor{blue!15} 0.156  
 & \cellcolor{blue!15} 0.296 & \cellcolor{blue!15} 0.032 & \cellcolor{blue!15} 0.108 & \cellcolor{blue!15} 0.338\\
 & \multirow{2}{*}{blotchy} & TCAV  
 & 0.459 & 0.210 & 0.457 & 1.460
 & 0.574 & 0.107 & 0.186 & 0.871 
 & 0.438 & 0.133 & 0.304 & 1.005\\
 &  & \cellcolor{blue!15} TGCAV  
 & \cellcolor{blue!15} 0.393 & \cellcolor{blue!15} 0.057 & \cellcolor{blue!15} 0.144 & \cellcolor{blue!15} 0.508  
 & \cellcolor{blue!15} 0.618 & \cellcolor{blue!15} 0.026 & \cellcolor{blue!15} 0.042 & \cellcolor{blue!15} 0.129 
 & \cellcolor{blue!15} 0.386 & \cellcolor{blue!15} 0.043 & \cellcolor{blue!15} 0.111 & \cellcolor{blue!15} 0.466\\
\midrule
\multirow{6}{*}{Zebra} 
 & \multirow{2}{*}{dotted} & TCAV  
 & 0.463 & 0.218 & 0.470 & 1.576  
 & 0.470 & 0.063 & 0.135 & 0.426
 & 0.484 & 0.130 & 0.268 & 0.950 \\
 &  & \cellcolor{blue!15} TGCAV  
 & \cellcolor{blue!15} 0.373  & \cellcolor{blue!15}  0.019 & \cellcolor{blue!15} 0.051 & \cellcolor{blue!15} 0.161  
 & \cellcolor{blue!15} 0.479 & \cellcolor{blue!15} 0.046 & \cellcolor{blue!15} 0.096 & \cellcolor{blue!15} 0.376 
 & \cellcolor{blue!15} 0.408 & \cellcolor{blue!15} 0.021 & \cellcolor{blue!15} 0.051 & \cellcolor{blue!15} 0.196\\
 & \multirow{2}{*}{striped} & TCAV  
 & 0.783 & 0.185 & 0.236 & 0.664  
 & 0.593 & 0.073 & 0.123 & 0.421
 & 0.601 & 0.145 & 0.241 & 0.832\\
 &  & \cellcolor{blue!15} TGCAV  
 & \cellcolor{blue!15} 0.699 & \cellcolor{blue!15}  0.042 & \cellcolor{blue!15} 0.060 & \cellcolor{blue!15} 0.200  
 & \cellcolor{blue!15} 0.597 & \cellcolor{blue!15} 0.050 & \cellcolor{blue!15} 0.084 & \cellcolor{blue!15} 0.318 
 & \cellcolor{blue!15} 0.600 & \cellcolor{blue!15} 0.017 & \cellcolor{blue!15} 0.029 & \cellcolor{blue!15} 0.083\\
 & \multirow{2}{*}{zigzagged} & TCAV  
 & 0.647 & 0.232 & 0.359 & 0.974  
 & 0.578 & 0.051 & 0.088 & 0.277 
 & 0.618 & 0.095 & 0.153 & 0.615\\
 &  & \cellcolor{blue!15} TGCAV  
 & \cellcolor{blue!15} 0.503 & \cellcolor{blue!15} 0.027 & \cellcolor{blue!15} 0.055 & \cellcolor{blue!15} 0.159  
 & \cellcolor{blue!15} 0.597 & \cellcolor{blue!15} 0.049 & \cellcolor{blue!15} 0.083 & \cellcolor{blue!15} 0.268 
 & \cellcolor{blue!15} 0.628 & \cellcolor{blue!15} 0.031 & \cellcolor{blue!15} 0.049 & \cellcolor{blue!15} 0.175\\
\midrule
\multirow{6}{*}{\makecell{Honey\\-comb}} 
 & \multirow{2}{*}{\makecell[l]{honey\\-combed}} & TCAV  
 & 0.836 & 0.132 & 0.158 & 0.419  
 & 0.648 & 0.135 & 0.208 & 0.818 
 & 0.572 & 0.106 & 0.186 & 0.612\\
 &  & \cellcolor{blue!15} TGCAV  
 & \cellcolor{blue!15} 0.897 & \cellcolor{blue!15} 0.100 & \cellcolor{blue!15} 0.111 & \cellcolor{blue!15} 0.323  
 & \cellcolor{blue!15} 0.591 & \cellcolor{blue!15} 0.046 & \cellcolor{blue!15} 0.078 & \cellcolor{blue!15} 0.271 
 & \cellcolor{blue!15} 0.511 & \cellcolor{blue!15} 0.018 & \cellcolor{blue!15} 0.034 & \cellcolor{blue!15} 0.098\\
 & \multirow{2}{*}{stratified} & TCAV  
 & 0.579 & 0.164 & 0.283 & 0.846  
 & 0.519 & 0.103 & 0.198 & 0.905
 & 0.474 & 0.066 & 0.140 & 0.464\\
 &  & \cellcolor{blue!15} TGCAV  
 & \cellcolor{blue!15} 0.574 & \cellcolor{blue!15} 0.059 & \cellcolor{blue!15} 0.103 & \cellcolor{blue!15} 0.331  
 & \cellcolor{blue!15} 0.473 & \cellcolor{blue!15} 0.042 & \cellcolor{blue!15} 0.089 & \cellcolor{blue!15} 0.254 
 & \cellcolor{blue!15} 0.508 & \cellcolor{blue!15} 0.017 & \cellcolor{blue!15} 0.033 & \cellcolor{blue!15} 0.098\\
 & \multirow{2}{*}{paisley} & TCAV  
 & 0.464 & 0.106 & 0.229 & 0.840  
 & 0.374 & 0.104 & 0.278 & 1.124
 & 0.441 & 0.072 & 0.162 & 0.567\\
 &  & \cellcolor{blue!15} TGCAV  
 & \cellcolor{blue!15} 0.448 & \cellcolor{blue!15} 0.029 & \cellcolor{blue!15} 0.065 & \cellcolor{blue!15} 0.223  
 & \cellcolor{blue!15} 0.311 & \cellcolor{blue!15} 0.037 & \cellcolor{blue!15} 0.118 & \cellcolor{blue!15} 0.321
 & \cellcolor{blue!15} 0.460 & \cellcolor{blue!15} 0.035 & \cellcolor{blue!15} 0.075 & \cellcolor{blue!15} 0.283 \\
\bottomrule
\end{tabularx}
\caption{Statistic results for different models (GoogleNet, ResNet50V2, MobileNetV2). 
Mean represents the average TCAV/TGCAV scores across layers, indicating concept importance. 
Std (Standard Deviation) measures score variability, where lower values imply greater stability. 
CV (Coefficient of Variation) quantifies relative dispersion, with lower values indicating higher consistency. 
RR (Range Ratio) captures score fluctuations across layers, where a smaller RR suggests more consistent scores across layers.}
\label{tab:statistic_result}
\end{table*}

To quantitatively evaluate the effectiveness of our method, we compute several statistical metrics over the TCAV scores, as shown in Table~\ref{tab:statistic_result}. We consider the following key metrics:

\begin{itemize}
    \item \textbf{Mean}: The average TCAV/TGCAV score across layers, indicating overall concept attribution.  
    \item \textbf{Standard Deviation (Std)}: Quantifies score dispersion across layers; lower values indicate greater stability.  
    \item \textbf{Coefficient of Variation (CV)}: The normalized variability of scores, defined as \(\text{CV} = \frac{\text{Std}}{\text{Mean}}\), facilitating comparison across concepts.  
    \item \textbf{Range Ratio (RR)}: Measures relative score fluctuation across layers, given by  
    \begin{equation}
    RR = \frac{\max(S) - \min(S)}{\operatorname{mean}(S)}
    \end{equation}
    where \( S \) denotes TCAV/TGCAV scores. Lower values indicate more consistent concept importance.  
\end{itemize}


From Table~\ref{tab:statistic_result}, we observe that our method significantly enhances the stability and reliability of TGCAV scores across layers. Across all models, the standard deviation, coefficient of variation (CV), interquartile range (IQR), and range ratio (RR) are consistently lower, indicating that TGCAV effectively reduces variance and stabilizes concept importance attribution. 

Taking the \emph{dotted} concept for \emph{Zebra} as an example, the standard deviation drops sharply in all three models, with GoogleNet showing the most dramatic reduction (0.218 to 0.019), while ResNet50V2 (0.063 to 0.046) and MobileNetV2 (0.130 to 0.021) also exhibit significant improvements. Notably, the coefficient of variation (CV) follows a similar trend, highlighting that the reduction in variability is not merely an effect of lower mean scores but rather a genuine stabilization of concept importance across layers. This trend is also evident in the \emph{zigzagged} concept, where the range ratio (RR) decreases notably, ensuring that TCAV scores are more evenly distributed across layers rather than dominated by a single one. This effect is particularly strong in GoogleNet, where the range ratio drops from 0.974 to 0.159. The consistently lower CV and RR values across models further confirm that TGCAV mitigates the layer-wise variability present in standard TCAV. 

At the same time, we find that the mean TGCAV scores remain largely aligned with their TCAV counterparts within each model, preserving the relative importance of concepts. However, the absolute mean values vary between models. For instance, in the \emph{honeycombed} concept for \emph{Honeycomb}, GoogleNet’s mean remains relatively stable (0.836 to 0.897), whereas ResNet50V2 (0.648 to 0.591) and MobileNetV2 (0.572 to 0.511) exhibit more noticeable shifts. These differences suggest that while TGCAV ensures consistency in concept attribution within each model, different architectures inherently emphasize different features. This variability likely reflects the models' differing internal representations and the features they prioritize when making predictions, contributing to their respective classification performances. The variations in CV values across models reinforce this point, as they suggest that some architectures assign concept importance more consistently than others.

\begin{figure}[h]
    \centering

       \includegraphics[width=\columnwidth]{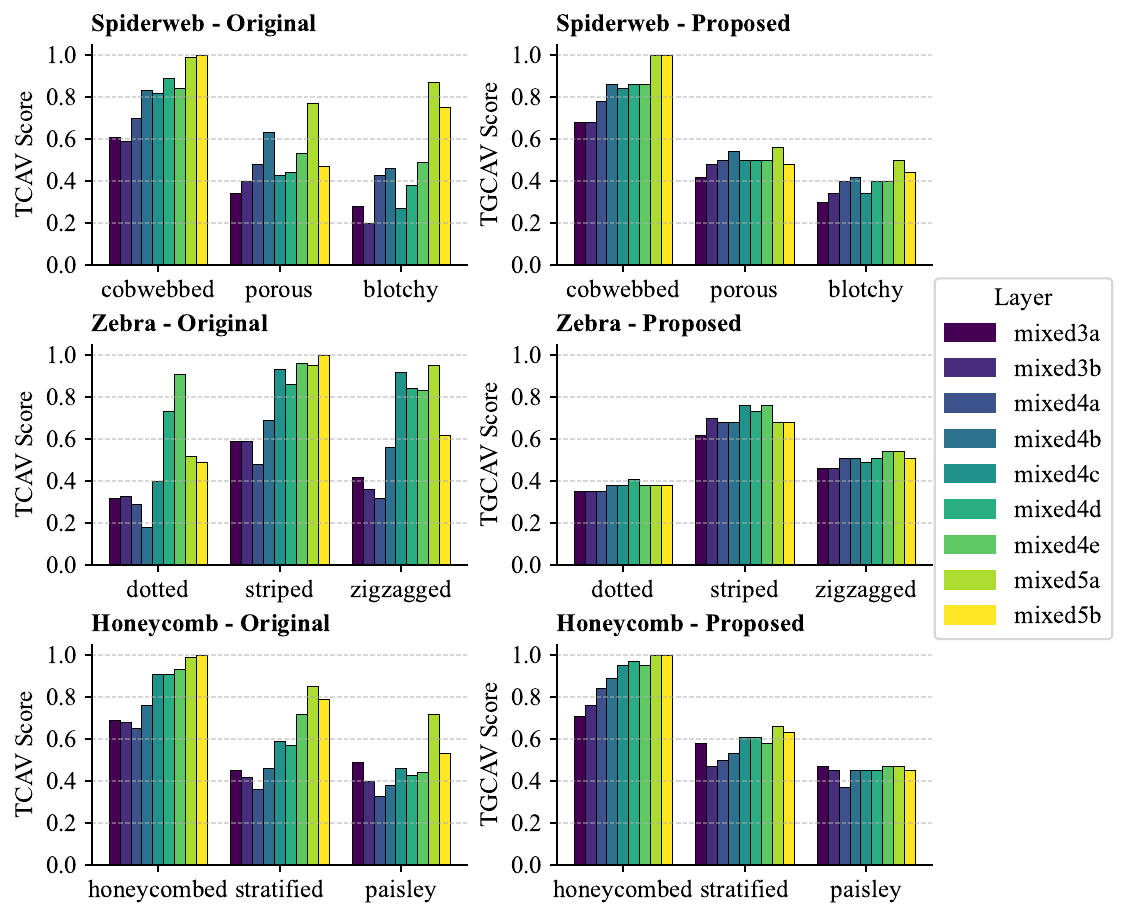}
    \caption{Comparison of TCAV scores across layers on GoogleNet before (left) and after (right) using our methods }
    \label{fig:tcav bar}
\end{figure}


Figure~\ref{fig:tcav bar} presents a layer-wise visualization of TCAV scores for different target classes (\textit{zebra, spider web, honeycomb}) before and after applying our method. The left column illustrates the TCAV scores computed using the original TCAV method, while the right column shows the results after incorporating our approach.

From the original TCAV results, we observe two major issues: (1) \textbf{High variance across layers}: The TCAV scores exhibit substantial fluctuations across different layers, making it unclear which layer should be used for interpretation. This instability is particularly evident in layers such as \texttt{mixed4e}, where all concepts receive uniformly high TCAV scores for the target class ``zebra", leading to unreliable concept attribution. (2) \textbf{Spurious high activations in irrelevant concepts}: Concepts such as \textit{dotted} and \textit{paisley} attain unexpectedly high scores in the \textit{zebra} and \textit{honeycomb} categories, respectively, despite being unrelated to the target classes. In contrast, the proposed method significantly stabilizes TGCAV scores across layers. The variance in TGCAV values is notably lower, and the relative importance of each concept remains consistent throughout the network. This ensures that meaningful concepts, such as \textit{striped} for \textit{zebra} and \textit{cobwebbed} for \textit{spider web}, maintain high TGCAV scores, while unrelated concepts no longer receive artificially inflated attributions. This improvement demonstrates the effectiveness of our method in enforcing \textbf{cross-layer consistency} while preserving the interpretability of concept importance. More visualization results can be seen in Appendix.

Together, these results confirm that our method effectively reduces TCAV’s dependence on layer selection while preserving the interpretability of concept importance.

\subsection{Concept Map Visualization for Precise Localization}

To further assess the effectiveness of our GCAV framework, we visualize the concept maps using the Visual-TCAV method~\cite{de2024visual} and compare them with those produced by the original approach. Unlike the original method, which computes CAVs independently for each layer, our method aligns and integrates concept representations across layers. To ensure a fair comparison, we aggregate concept maps from all layers into a single unified visualization.

\begin{figure*}[h]
    \centering
    \begin{subfigure}{0.33\textwidth}
        \centering
        \includegraphics[width=0.32\textwidth]{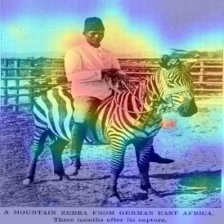}
        \includegraphics[width=0.32\textwidth]{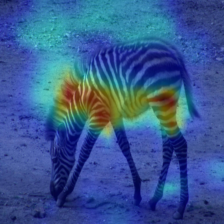}
        \includegraphics[width=0.32\textwidth]{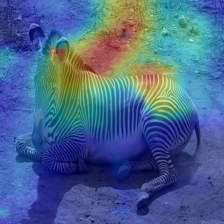}
        \includegraphics[width=0.32\textwidth]{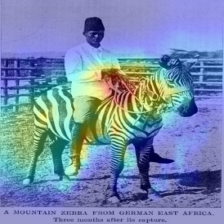}
        \includegraphics[width=0.32\textwidth]{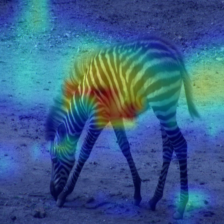}
        \includegraphics[width=0.32\textwidth]{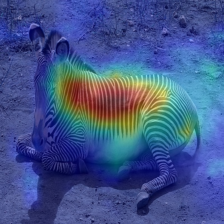}
        \caption{Concept Maps for the \textbf{striped} concept in the class ``zebra.''}
        \label{fig:concept_map_zebra}
    \end{subfigure}
        \begin{subfigure}{0.33\textwidth}
        \centering
        \includegraphics[width=0.32\textwidth]{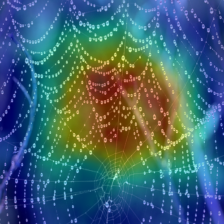}
        \includegraphics[width=0.32\textwidth]{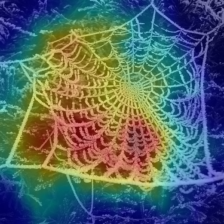}
        \includegraphics[width=0.32\textwidth]{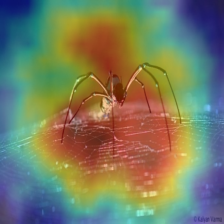}
        \includegraphics[width=0.32\textwidth]{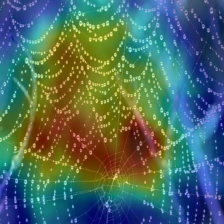}
        \includegraphics[width=0.32\textwidth]{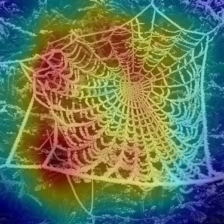}
        \includegraphics[width=0.32\textwidth]{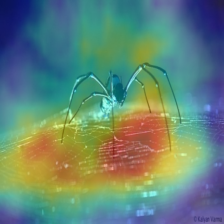}
        \caption{Concept Maps for the \textbf{cobwebbed} concept in the class ``spider web.''}
        \label{fig:concept_map_spider}
    \end{subfigure}
        \begin{subfigure}{0.33\textwidth}
        \centering
        \includegraphics[width=0.32\textwidth]{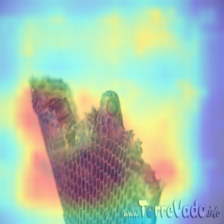}
        \includegraphics[width=0.32\textwidth]{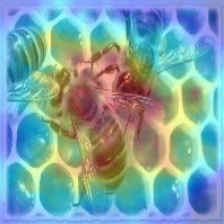}
        \includegraphics[width=0.32\textwidth]{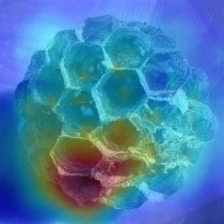}
        \includegraphics[width=0.32\textwidth]{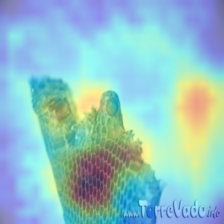}
        \includegraphics[width=0.32\textwidth]{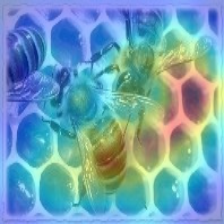}
        \includegraphics[width=0.32\textwidth]{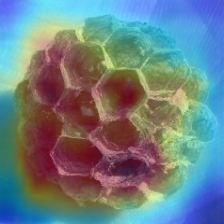}
        \caption{Concept Maps for the \textbf{honeycombed} concept in the class ``honeycomb.''}
        \label{fig:concept_map_honeycomb}
    \end{subfigure}
    \caption{Visual-TCAV Concept Maps of GoogleNet. The top row represents the original method, while the bottom row represents our proposed method. Our method focuses more precisely on the characteristic pattern while reducing distractions from irrelevant regions.}
    \label{fig:concept_maps}
\end{figure*}

Figure~\ref{fig:concept_maps} illustrates the concept maps for three representative concepts: \textit{striped} (zebra), \textit{cobwebbed} (spider web), and \textit{honeycombed} (honeycomb). The top row corresponds to the original TCAV method, while the bottom row presents the results of our proposed method.

The original method exhibits more dispersed activations, sometimes highlighting background regions or unrelated features. In contrast, our proposed method produces more concentrated activations, aligning better with the expected locations of each concept. For instance, in the \textit{striped} concept within the \textit{zebra} class (Figure~\ref{fig:concept_map_zebra}), the original method tends to activate regions beyond the zebra’s stripes, including parts of the sky and surrounding environment. Our approach, however, focuses more precisely on the zebra’s body, ensuring that the concept representation remains faithful to its intended meaning. A similar trend can be observed in the \textit{cobwebbed} and \textit{honeycombed} concepts (Figures~\ref{fig:concept_map_spider} and~\ref{fig:concept_map_honeycomb}), where the original method introduces unnecessary activations in non-characteristic areas, while our method maintains a sharper focus on the defining structural patterns.

These results suggest that our approach enhances TCAV’s interpretability by integrating global contextual information, which in turn enables more accurate semantic extraction and ultimately leads to more precise localization. This improvement ensures that the learned concepts are both spatially and semantically meaningful, providing more robust and consistent explanations in deep learning models.

\subsection{Case Study: Robustness Evaluation under Adversarial Attacks}
Since our GCAV method integrates information from all selected layers, it is inherently more robust to adversarial attacks that target specific layers. In this paper, we apply Brown's method~\cite{Brown2021MakingCI} to evaluate the robustness of GCAV against such attacks.

\begin{figure}[h]

    \centering
    \begin{subfigure}{0.23\textwidth}
        \centering
        \includegraphics[width=0.48\textwidth]{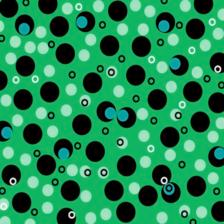}
        \includegraphics[width=0.48\textwidth]{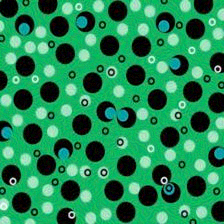}
    \end{subfigure}
    \begin{subfigure}{0.23\textwidth}
        \centering
        \includegraphics[width=0.48\textwidth]{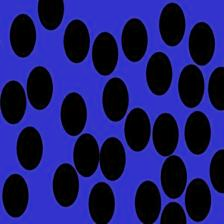}
        \includegraphics[width=0.48\textwidth]{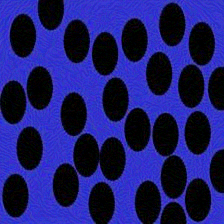}
    \end{subfigure}
   \begin{subfigure}{0.23\textwidth}
        \centering
        \includegraphics[width=0.48\textwidth]{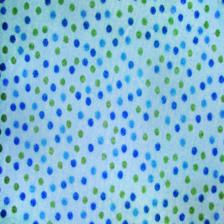}
        \includegraphics[width=0.48\textwidth]{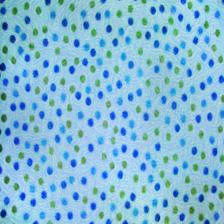}
    \end{subfigure}
    \begin{subfigure}{0.23\textwidth}
        \centering
        \includegraphics[width=0.48\textwidth]{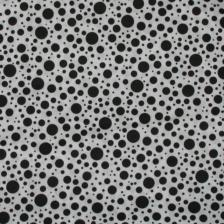}
        \includegraphics[width=0.48\textwidth]{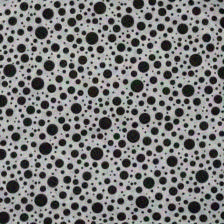}
    \end{subfigure}
    \caption{Adversarial Attack on the “dotted” Concept at GoogleNet \texttt{mixed5a} Layer. In each pair, the left image is the original concept, while the right image is the perturbed version disrupting the activation at \texttt{mixed5a}.}
    \label{fig:attack sample}
\end{figure}

To better illustrate the advantages of GCAV over traditional CAV-based methods, we present a case study focusing on the concept of ``dotted'' in the class ``zebra.'' We apply the \emph{Targeted Attack} proposed in~\cite{Brown2021MakingCI} to artificially increase its importance by shifting its activation values in the bottleneck layer \texttt{mixed5a} closer to those of the target concept ``striped.'' Figure~\ref{fig:attack sample} illustrates examples from the probe dataset after the attack.

As seen in Table \ref{tab:attacked results}, traditional TCAV experiences a substantial increase in TCAV scores under attack, with the mixed5a TCAV value for ``dotted" rising from 0.52 to 0.96, an increase of 84.61\%. This suggests that the attack successfully manipulates the interpretability results, making the concept appear significantly more relevant than it originally was. In contrast, GCAV demonstrates greater resilience, with the same attack causing only a 28.49\% increase, from 0.38 to 0.49. Similarly, when examining the mean TCAV scores across layers, TCAV shows a 61.77\% increase, whereas GCAV exhibits a smaller rise of 42.63\%, indicating that our method mitigates the extent of adversarial influence.
\begin{table}[ht]
\centering
\resizebox{\columnwidth}{!}{
\begin{tabular}{lcrcr}
\toprule
 Method & Mixed5a& Mixed5a Change(\%) & Mean &Mean Change (\%) \\
\midrule
TCAV& 0.52 &--& 0.46&-- \\
TCAV(Attacked)& 0.96& +84.61& 0.75& +61.77\\
TGCAV& 0.38 & -- &0.37& -- \\
TGCAV(Attacked)& 0.49& +28.49 & 0.53 & +42.63\\
\bottomrule
\end{tabular}}
\caption{Results of Adversarial Attack on \texttt{mixed5a} for the Concept ``dotted''}
\label{tab:attacked results}
\end{table}

The improved robustness of GCAV can be attributed to several key factors. First, our framework integrates information across multiple layers, reducing the impact of adversarial perturbations on any single layer. Unlike traditional TCAV, which relies on independently computed CAVs for each layer, our method aligns and fuses CAVs into a global representation, thereby preventing localized disruptions from significantly altering interpretability results. Second, our approach incorporates consistency constraints during training, ensuring that concept representations remain stable across layers. This enforces greater semantic coherence and prevents adversarially induced distortions in specific layers from propagating through the interpretability framework. Finally, the variance loss used in our fusion step further enhances robustness by minimizing the fluctuation of TGCAV scores across layers, making it harder for adversarial attacks to exploit weak points in the model.

These findings demonstrate that GCAV not only enhances interpretability consistency but also significantly improves robustness against adversarial manipulation. By leveraging a global concept representation rather than relying on isolated per-layer CAVs, our approach provides a more stable and reliable interpretability framework, making it more suitable for deployment in real-world scenarios where model explanations must be both trustworthy and resilient to adversarial threats.
\section{Conclusion}
\label{sec:Conclusion}

This paper introduces the Global Concept Activation Vector (GCAV), a framework that unifies Concept Activation Vectors across layers to enhance interpretability consistency. By integrating contrastive learning for alignment and an attention-based fusion mechanism, GCAV constructs a stable, globally consistent representation. We further propose Testing with Global Concept Activation Vectors (TGCAV) to apply this unified concept representation in interpretability analysis.

Experimental results show that TGCAV significantly reduces layer-wise variability, leading to more stable concept attributions across different architectures. Concept map visualizations demonstrate improved concept localization, while robustness evaluations confirm that GCAV mitigates adversarial vulnerabilities present in standard TCAV. Additionally, our findings reveal that different models prioritize concepts differently, reflecting their distinct internal representations.

Future work could explore leveraging GCAV for causal analysis of layer interactions, investigating how concepts propagate across layers and influence model decisions. This could provide deeper insights into the hierarchical structure of deep networks and improve the reliability of concept-based explanations.

\section*{Acknowledgements}
This work is supported in part by the US National Science Foundation under grants 2217071, 2213700, 2106913, 2008208.

{
    \small
    \bibliographystyle{ieeenat_fullname}
    \bibliography{main}
}

\clearpage
\onecolumn
\appendix
\label{sec:appendix}
\section{Implementation Details}
All experiments are conducted on a high-performance computing cluster equipped with four NVIDIA A100 80GB GPUs. The system runs on CUDA 12.4 with NVIDIA driver version 550.54.14. The deep learning framework used for implementation is PyTorch, and model training is optimized using mixed precision and multi-GPU distributed training to maximize efficiency.

\subsection{Dimension Unification}

The encoder and decoder are designed symmetrically, each consisting of two fully connected layers. For GoogleNet and MobileNetV2, the hidden layer has 4096 units, and the final embedding layer also has 4096 units. For ResNet50V2, both the hidden and embedding layers are reduced to 2048 units to optimize GPU memory usage and prevent out-of-memory (OOM) errors during training. Each layer is followed by a ReLU activation function to introduce non-linearity and improve representation learning.

\subsection{Contrastive Learning for CAV Alignment}
\label{appendix:align}

\subsubsection{MLP Projection Head design}
Here we introduce the details of implementation of the MLP Projection head in Figeure~\ref{framework}.
To ensure that embeddings obtained from different layers are projected into a shared semantic space, we employ a residual MLP architecture for \( f(\cdot) \). This design follows two key principles:

\begin{itemize}
    \item \textbf{Alignment Across Layers:} Since embeddings come from different autoencoders trained on separate layers, their distributions may vary. The MLP projection helps unify these representations.
    \item \textbf{Preserving Original Semantics:} A residual connection ensures that the transformed embeddings retain the essential concept-specific information from the original embeddings.
\end{itemize}

The function is defined as follows:

\begin{verbatim}
def projection_function(z):
    # Non-linear transformation through MLP
    projected = MLP(z)  
    # Residual connection with a linear transformation
    residual = Linear(z)
    # Normalization to maintain consistent scale
    return normalize(projected + residual)
\end{verbatim}

Here, ``MLP(z)" denotes a multi-layer perceptron with LayerNorm and GELU activations, ensuring smooth and stable training. ``Linear(z)" represents a simple linear layer that directly maps the input to the output space, enabling residual learning. The final normalization step ensures that the projected embeddings remain in a comparable range.

This projection function plays a crucial role in learning a unified representation for concept activation vectors across different layers.




\subsection{Cross-Layer CAV Fusion}
\label{appendix:fuse}

In this section, we detail the design of our fusion module and the loss function used to train the model.

\subsubsection{Transformer-Based Cross-Layer CAV Fusion}

To fuse CAV embeddings from different layers, we employ an attention-based fusion module. Given a set of aligned CAV embeddings 
\[
\{\mathbf{z}_l\}_{l=1}^{L}, \quad \mathbf{z}_l \in \mathbb{R}^{d},
\]
the fusion module computes a global CAV representation. In particular, positional encoding is added to the input embeddings, followed by a multi-head self-attention layer and a feed-forward network (FFN). A residual connection and layer normalization are applied at each stage. Finally, an average pooling across the layer dimension yields the global CAV:
\begin{equation}
\mathbf{z}_{\text{global}} = f_{\text{fuse}}(\{\mathbf{z}_l\}) = \text{OutputLayer}\left(\frac{1}{L}\sum_{l=1}^{L}\tilde{\mathbf{z}}_l\right).
\end{equation}
The pseudo-code for the fusion module is as follows:

\begin{verbatim}
Input: CAV embeddings [batch_size, num_layers, embedding_dim]
1. Add positional encoding to each layer's embedding.
2. Compute self-attention via a multi-head attention layer.
3. Apply residual connection and layer normalization.
4. Process the result with a feed-forward network and dropout.
5. Apply a second residual connection and normalization.
6. Aggregate across layers (e.g., average pooling).
7. Transform the pooled result via a linear output layer.
Output: Global CAV [batch_size, embedding_dim]
\end{verbatim}

\subsubsection{Loss Design}

The overall loss function consists of two components:  

\begin{itemize}
    \item \(\mathcal{L}_{\text{var}}\): The \textbf{layer variance loss}, which enforces layer-wise consistency by minimizing the variance of TCAV scores across layers.
    \item \(\mathcal{L}_{\text{cons}}\): The \textbf{concept center separation loss}, which ensures that the reconstructed CAVs remain semantically aligned with the original ones by enforcing a cosine-similarity constraint.
\end{itemize}

The final loss function is formulated as:
\begin{equation}
\mathcal{L} = \lambda_{\text{var}} \, \mathcal{L}_{\text{var}} + \lambda_{\text{cons}} \, \mathcal{L}_{\text{cons}},
\end{equation}
where \(\lambda_{\text{var}}\) and \(\lambda_{\text{cons}}\) are balancing weights that regulate the trade-off between cross-layer consistency and semantic preservation.

The loss computation follows these steps:

\begin{verbatim}
Input: 
    - GCAV [batch_size, cav_dim]
    - Decoders for each layer
    - Class activation values per layer
    - Class examples and target labels
    - Additional model parameters (e.g., bottleneck, concept labels)
    
1. Update the dynamic temperature parameter.
2. For each layer:
    a. Reconstruct the CAV using the corresponding decoder.
    b. Compute cosine similarity between the reconstructed and original CAV.
    c. Compute the TCAV score using a gradient-based method.
3. Compute consistency loss as the average cosine loss over layers.
4. Compute variance loss as the variance of TCAV scores across layers.
5. Compute total loss:
       total_loss = var_weight * variance_loss + consistency_weight * consistency_loss.
6. Return total_loss.
\end{verbatim}

These components ensure that the fusion module aggregates multi-layer information into a unified GCAV while preserving consistency and maintaining clear semantic separation between concepts.

During training, the computation of TCAV scores \(s_l\) at layer \( l \) follows the standard TCAV formulation. As described in Section~\ref{sec:Cross-Layer CAV Fusion}, we obtain the reconstructed concept vector \( \tilde{\mathbf{v}_l^c} \) by decoding the globally integrated representation \( \mathbf{z}_{GCAV}^c \) using the layer-specific decoder\(f_{\text{decoder}}^l\). We then replace the original CAV with \( \tilde{\mathbf{v}_l^c} \) and compute the TCAV score as:

\begin{equation}
    TCAV_{c, k,l} = \frac{\big|\{ x \in X_k : \nabla h_{l,k}\bigl(f_l(x)\bigr) \cdot \tilde{\mathbf{v}_l^c}  > 0 \} \big|}{|X_k|},
\end{equation}
where \(X_k\) denotes the set of inputs belonging to class \(k\), \( f_l(x) \) represents the activation at layer \( l \) for input \( x \), and \( h_{l,k} : \mathbb{R}^m \to \mathbb{R} \) maps these activations to the logit for class \( k \). This formulation measures the proportion of class examples for which the directional derivative along \( \tilde{\mathbf{v}_l^c} \) aligns with the classification objective.

To enable gradient-based optimization, we approximate the non-differentiable "$>$" operator using a sigmoid relaxation:

\begin{equation}
    s^{(i)}_l = \operatorname{STE}\Bigl(\sigma\Bigl(-\tau\, \nabla h_{l,k}(f_l(x)) \cdot \tilde{\mathbf{v}_l^c}\Bigr)\Bigr),
\end{equation}
where \( \sigma(x) = \frac{1}{1 + e^{-x}} \) is the sigmoid function, \( \operatorname{STE} \) is the straight-through estimator~\cite{bengio2013estimating}, and \( \tau \) is a temperature parameter that gradually increases during training. This progressively sharpens the approximation of the hard threshold, allowing the loss to remain differentiable while converging toward the standard TCAV formulation.

By integrating these loss components, the model learns a globally aligned concept representation that remains stable across layers while preserving the distinctiveness of individual concepts.

\section{More Results}
\label{sec:more results}
This section provides ablation studies, hyperparameter investigations, and extensive tests on high-level concepts using more images from additional datasets. We also present results from additional visualizations of TCAV and TGCAV scores across layers in different models, including GoogleNet, MobileNetV2, and ResNet50V2. 

\subsection{Ablation Study}  
Table~\ref{tab:ablation} shows ablations for the “Zebra” class on GoogleNet. “w/o Align” and “w/o Fuse” remove the alignment or fusion step, respectively. This complements Table 1 in our paper. Removing align increases variance, while removing fusion inflates all scores, indicating that fusion integrates multi-layer signals to reduce noise and improve consistency.

\begin{table}[h]
\centering
\begin{tabular}{lcccccccc}
\toprule
\textbf{Concept} &
\multicolumn{4}{c}{\textbf{w/o Align}} &
\multicolumn{4}{c}{\textbf{w/o Fuse}} \\
\cmidrule(lr){2-5} \cmidrule(lr){6-9}
& Mean & Std & CV & RR & Mean & Std & CV & RR \\
\midrule
dotted    & 0.299 & 0.052 & 0.175 & 0.468 & 0.614 & 0.297 & 0.484 & 1.237 \\
striped   & 0.844 & 0.134 & 0.159 & 0.426 & 0.672 & 0.220 & 0.327 & 0.952 \\
zigzagged & 0.613 & 0.139 & 0.227 & 0.587 & 0.627 & 0.281 & 0.448 & 1.197 \\
\bottomrule
\end{tabular}
\caption{Ablation Study  for the “Zebra” class on GoogleNet.}
\label{tab:ablation}
\end{table}

\subsection{Hyperparameter Investigation}
\noindent Table~\ref{tab:hyper} shows TGCAV scores for \textit{striped} on \textit{zebra} using GoogleNet. We vary one setting at a time, fixing others as default ($\lambda_{\text{var}}{:}\lambda_{\text{cons}}=3{:}1$, $\lambda_{\text{NCE}}{:}\lambda_{\text{cons}}=1{:}3$, embedding = 4096), which are used in our main paper. We tested the case with no consistency loss ($\lambda_{\text{cons}} = 0$ in both stage). This caused TGCAV scores to collapse to $\sim$0.5 across layers, discarding meaningful concept signals. This motivated its inclusion in the final design. Embedding size showed minor effect; we selected a higher value (4096) for expressiveness.

\begin{table}[h]
\centering
\begin{tabular}{lcccccc}
\toprule
\textbf{Hyperparameter} & \textbf{Value} & \textbf{Mean} & \textbf{Std} & \textbf{CV} & \textbf{RR} \\
\midrule
$\lambda_{\text{var}} : \lambda_{\text{cons}}$ & 1 : 1  & 0.841 & 0.172 & 0.205 & 0.511 \\
$\lambda_{\text{var}} : \lambda_{\text{cons}}$ & 1 : 3  & 0.820 & 0.192 & 0.234 & 0.524 \\
$\lambda_{\text{NCE}} : \lambda_{\text{cons}}$ & 1 : 1  & 0.643 & 0.107 & 0.166 & 0.486 \\
$\lambda_{\text{NCE}} : \lambda_{\text{cons}}$ & 3 : 1  & 0.510 & 0.143 & 0.280 & 0.506 \\
\midrule
Embedding dim & 1024  & 0.701 & 0.042 & 0.059 & 0.112 \\
Embedding dim & 2048  & 0.750 & 0.069 & 0.092 & 0.107 \\
\bottomrule
\end{tabular}
\caption{Hyperparameter Investigation.}
\label{tab:hyper}
\end{table}

\subsection{Testing with High-level Concepts on More Datasets}
As shown in table~\ref{tab:more}, we evaluated three high-level concepts on CelebA, StanfordCars, and CUB-200-2011, using 500 images per concept. Our method still reduces inconsistency, showing good generalization.
\begin{table}[h]
\centering
\begin{tabular}{lcccccccc}
\toprule
\multicolumn{1}{c}{} & \multicolumn{4}{c}{\textbf{TCAV}} & \multicolumn{4}{c}{\textbf{TGCAV}} \\
\cmidrule(lr){2-5} \cmidrule(lr){6-9}
\textbf{Concept (Task)} & Mean & Std & CV & RR & Mean & Std & CV & RR \\
\midrule
human (groom) & 0.333 & 0.175 & 0.526 & 1.740 & 0.432 & 0.056 & 0.130 & 0.323 \\
car (cab)     & 0.608 & 0.176 & 0.289 & 0.888 & 0.692 & 0.098 & 0.142 & 0.103 \\
bird (robin)  & 0.623 & 0.132 & 0.212 & 0.674 & 0.639 & 0.102 & 0.160 & 0.496 \\
\bottomrule
\end{tabular}
\caption{Testing with High-level Concepts on CelebA, StanfordCars, and CUB-200-2011.}
\label{tab:more}
\end{table}

\subsection{Bar Chart Analysis}
\begin{figure}[h]
    \centering
    \begin{subfigure}{0.33\textwidth}
        \centering
        \includegraphics[width=\textwidth]{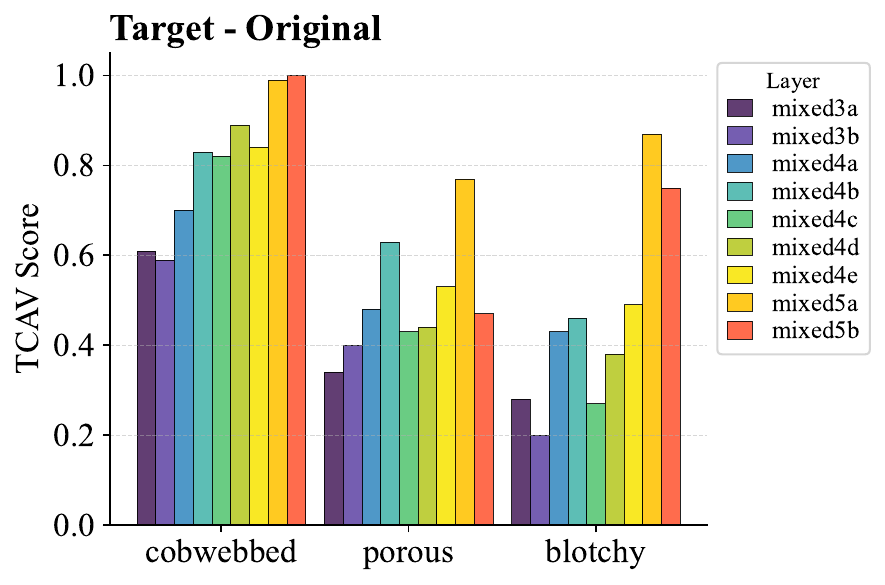}
        \includegraphics[width=\textwidth]{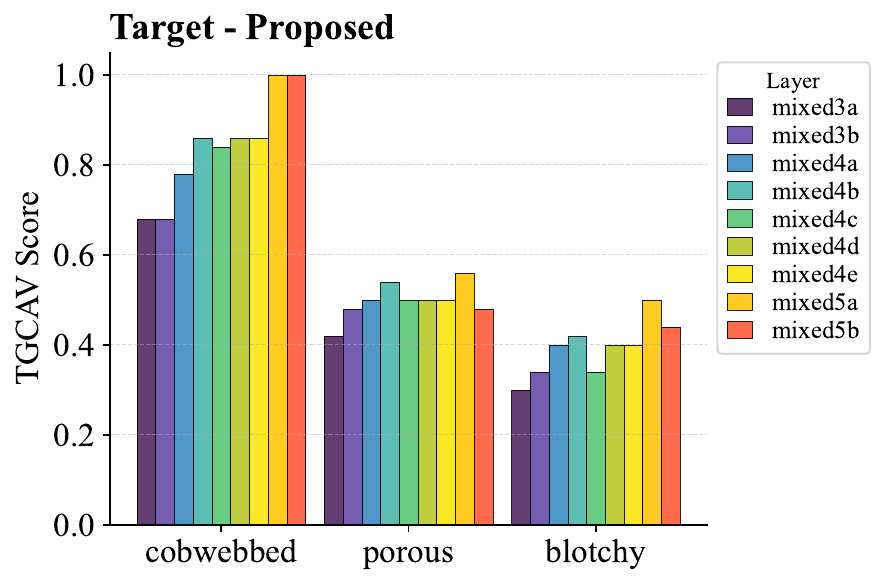}
    \end{subfigure}
    \begin{subfigure}{0.33\textwidth}
        \centering
        \includegraphics[width=\textwidth]{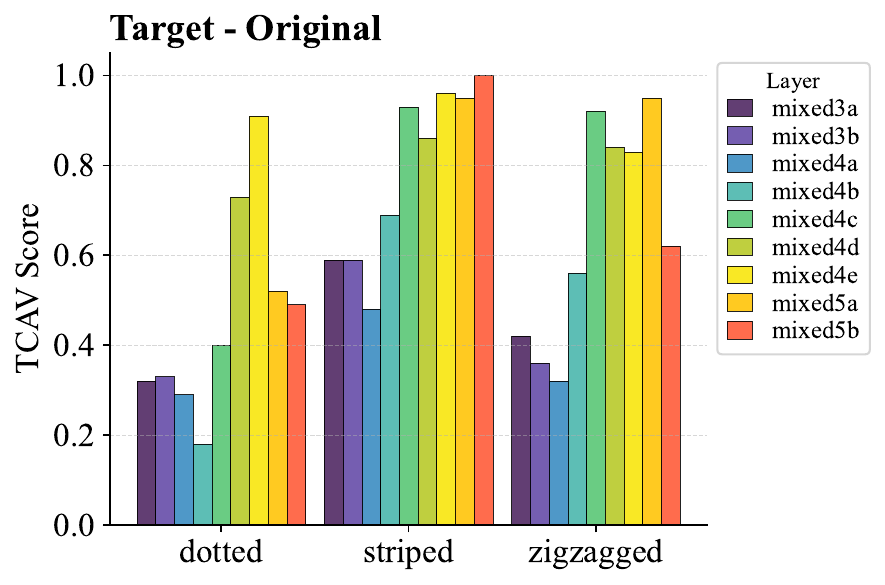}
        \includegraphics[width=\textwidth]{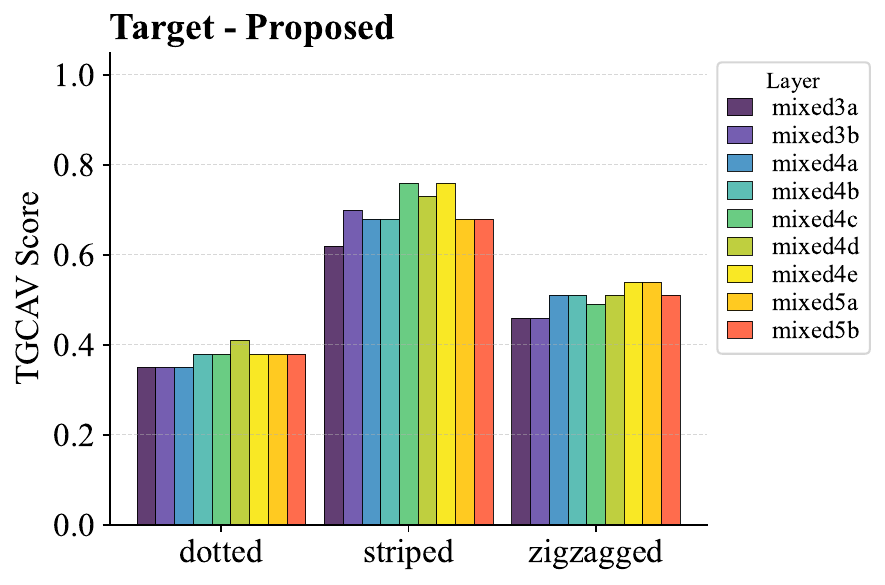}
    \end{subfigure}
        \begin{subfigure}{0.33\textwidth}
        \centering
        \includegraphics[width=\textwidth]{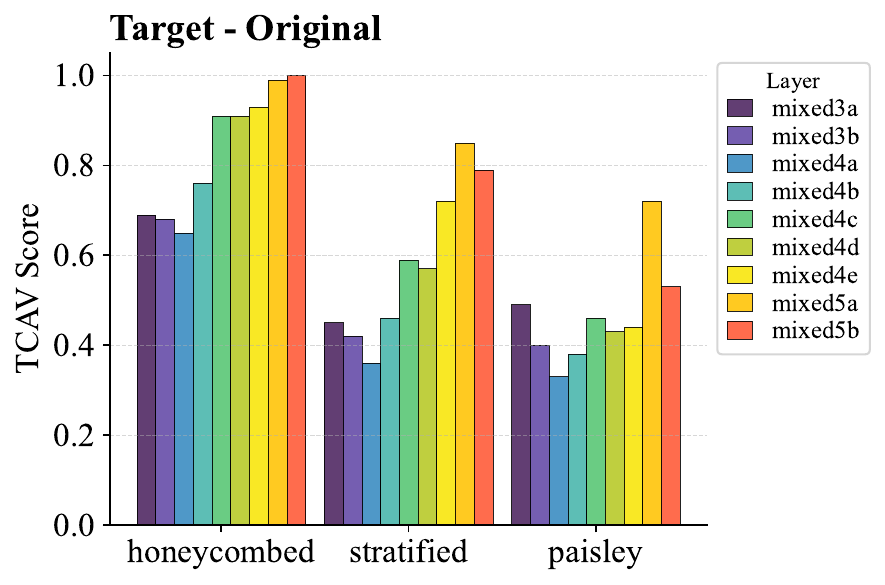}
        \includegraphics[width=\textwidth]{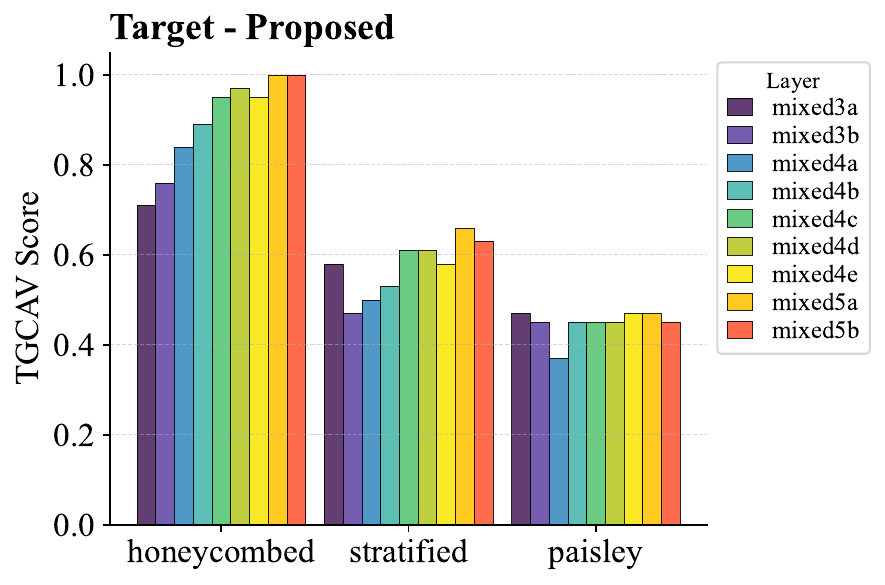}
    \end{subfigure}
    \caption{Bar Chart of TCAV scores of GoogleNet.}
    \label{fig:tcav bar googlenet}
\end{figure}

\begin{figure}[h]
    \centering
    \begin{subfigure}{0.33\textwidth}
        \centering
        \includegraphics[width=\textwidth]{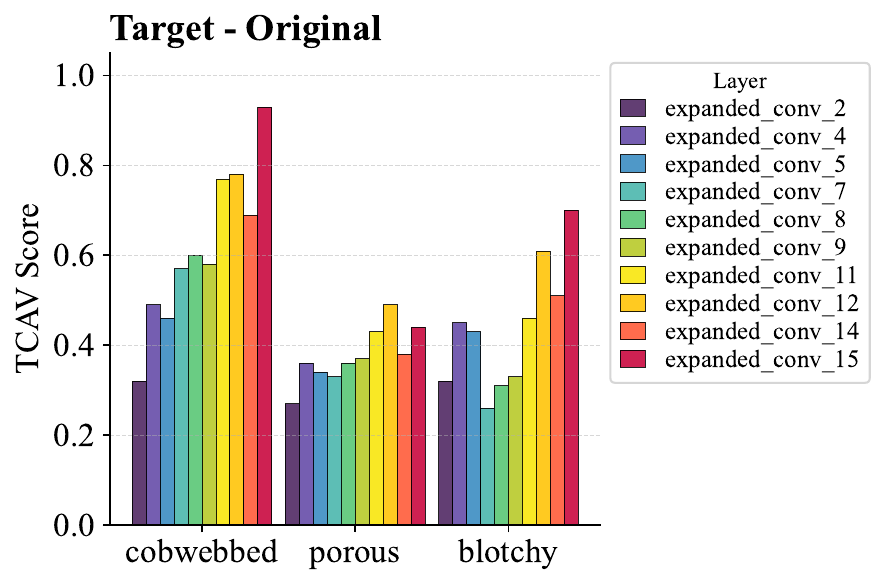}
        \includegraphics[width=\textwidth]{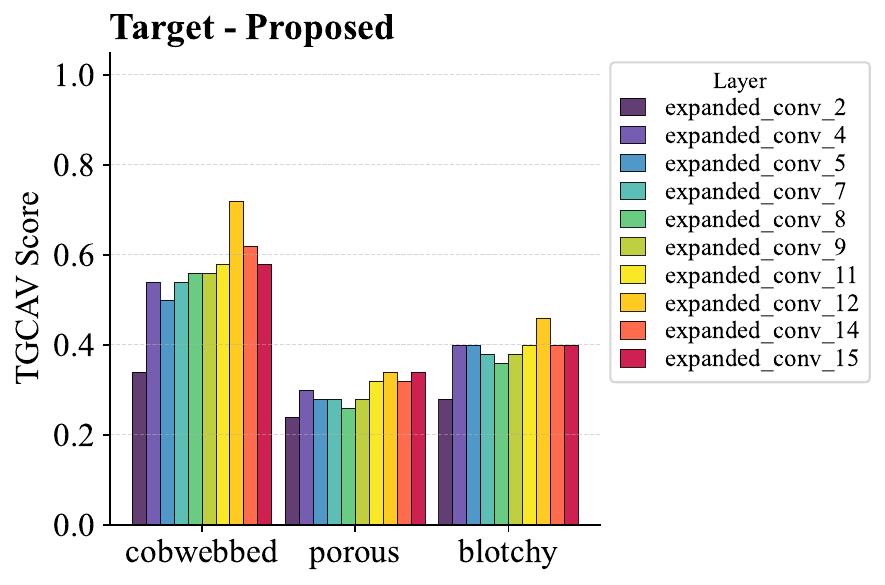}
    \end{subfigure}
    \begin{subfigure}{0.33\textwidth}
        \centering
        \includegraphics[width=\textwidth]{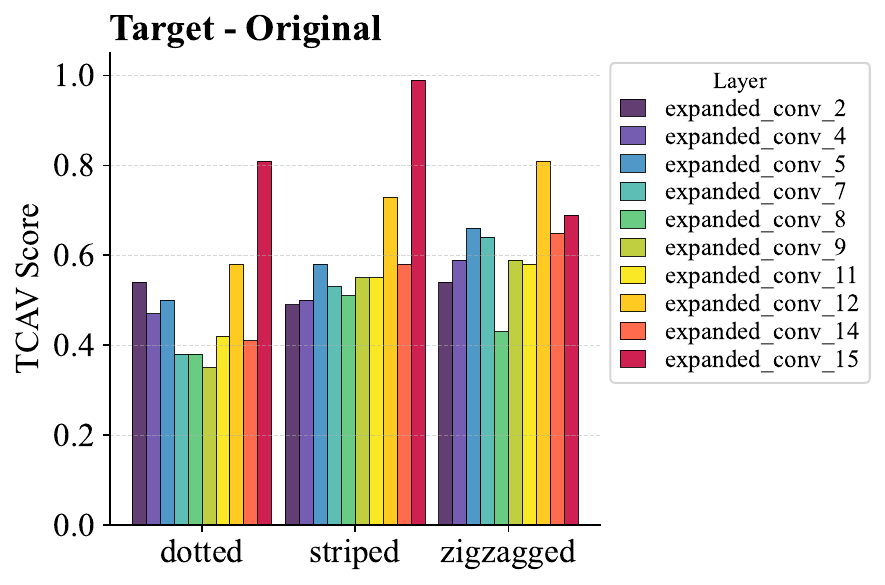}
        \includegraphics[width=\textwidth]{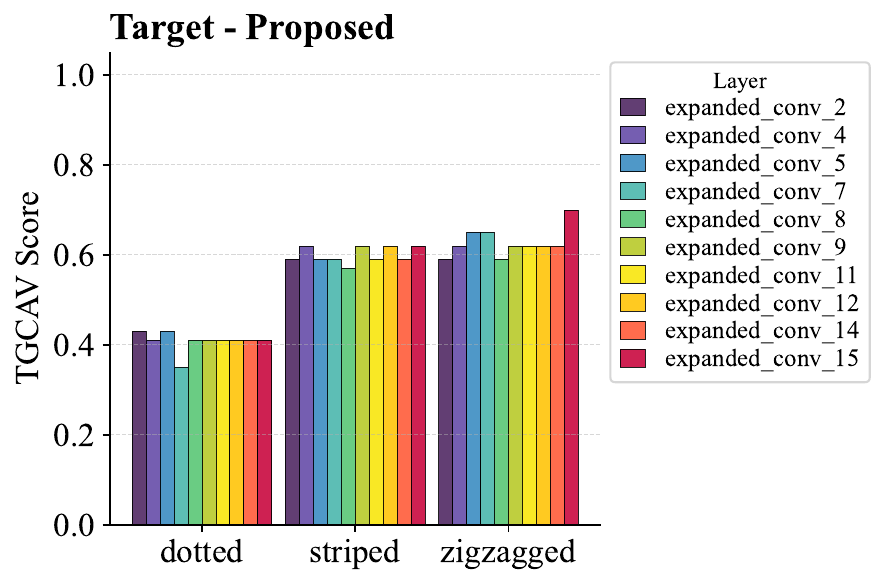}
    \end{subfigure}
        \begin{subfigure}{0.33\textwidth}
        \centering
        \includegraphics[width=\textwidth]{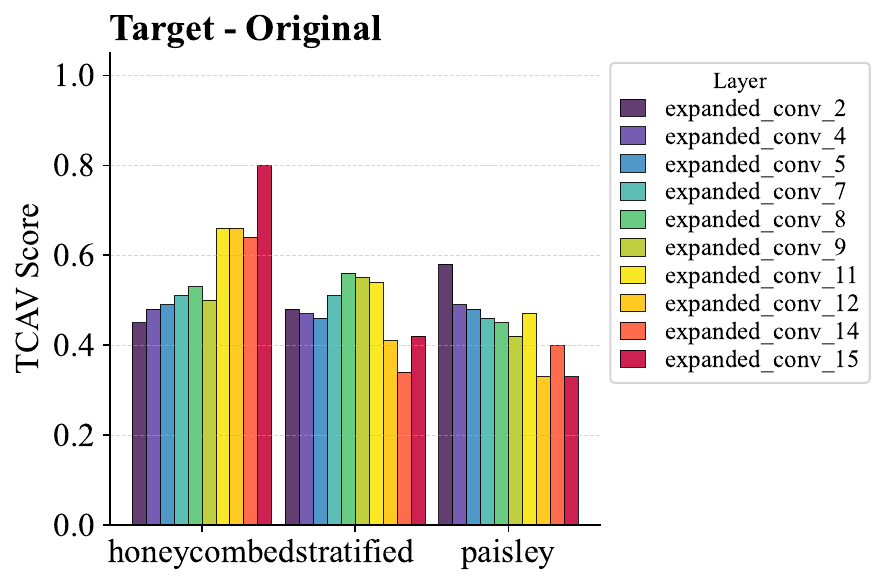}
        \includegraphics[width=\textwidth]{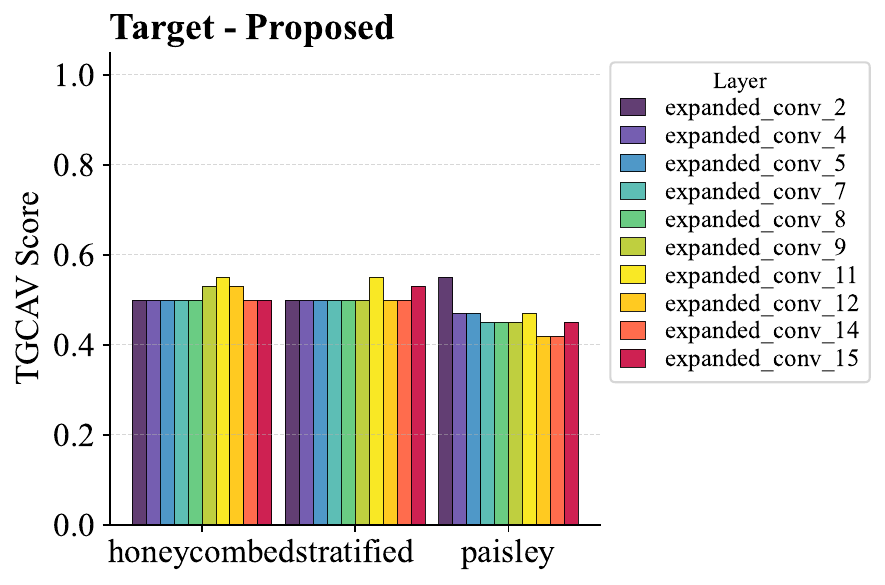}
    \end{subfigure}
    \caption{Bar Chart of TCAV scores of MobileNetV2.}
    \label{fig:tcav bar MobileNetV2}
\end{figure}

\begin{figure}[h]
    \centering
    \begin{subfigure}{0.33\textwidth}
        \centering
        \includegraphics[width=\textwidth]{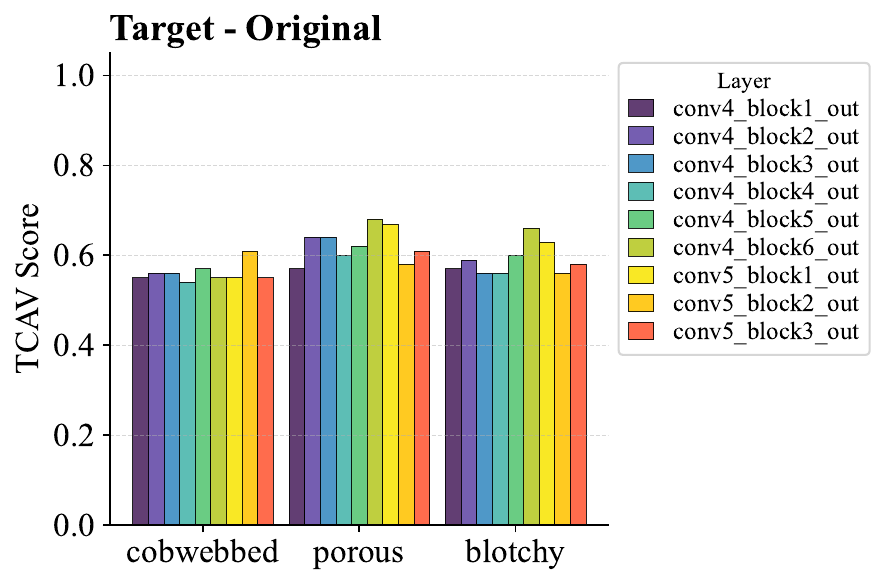}
        \includegraphics[width=\textwidth]{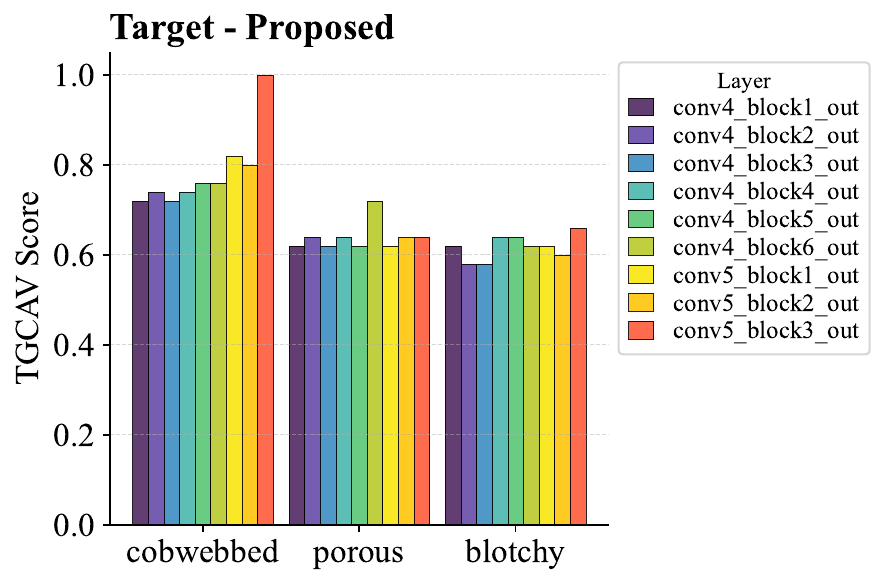}
    \end{subfigure}
    \begin{subfigure}{0.33\textwidth}
        \centering
        \includegraphics[width=\textwidth]{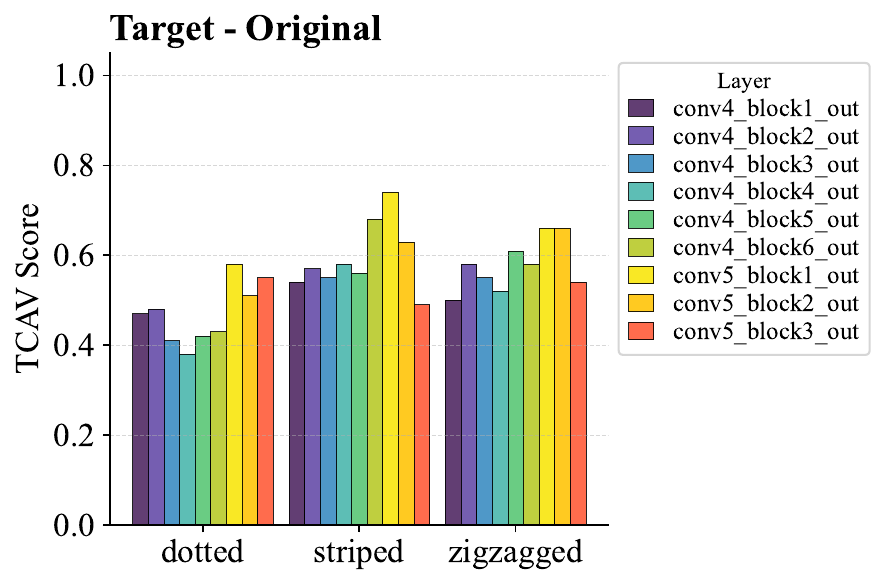}
        \includegraphics[width=\textwidth]{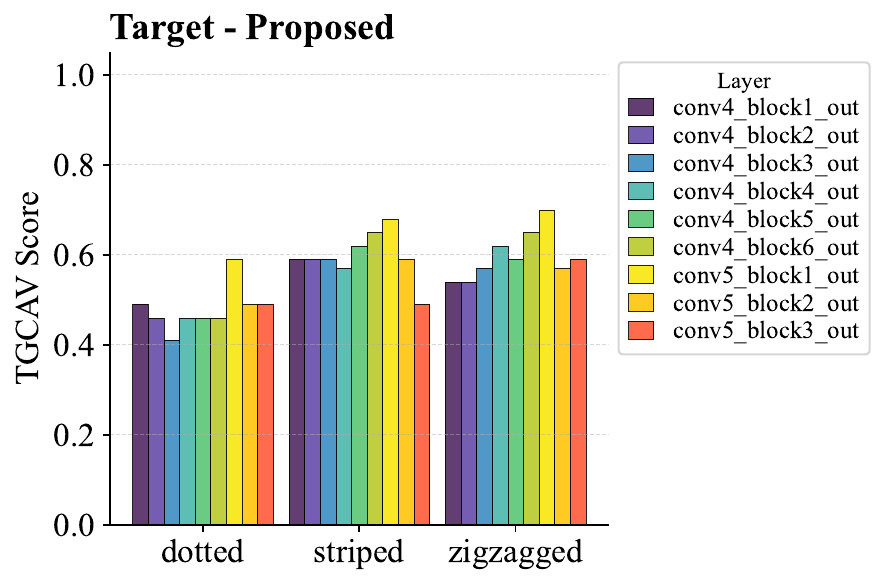}
    \end{subfigure}
        \begin{subfigure}{0.33\textwidth}
        \centering
        \includegraphics[width=\textwidth]{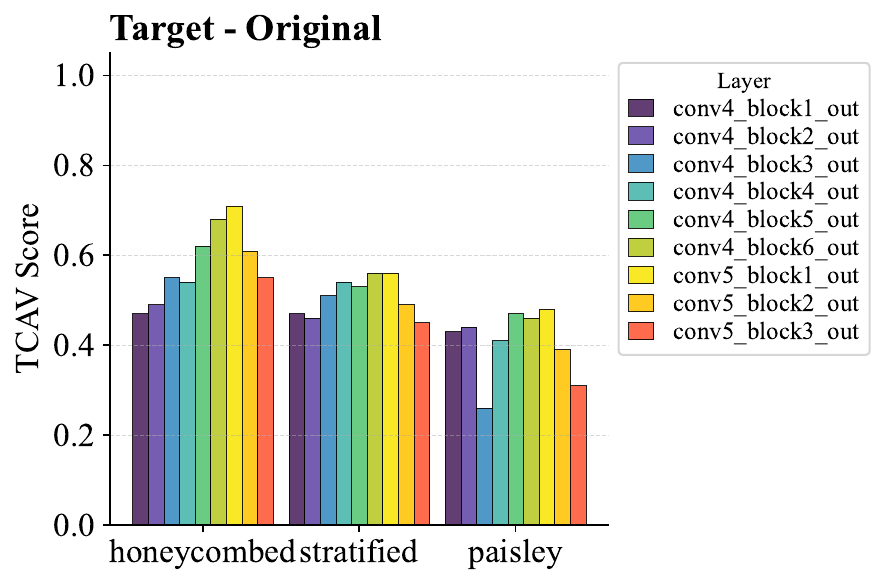}
        \includegraphics[width=\textwidth]{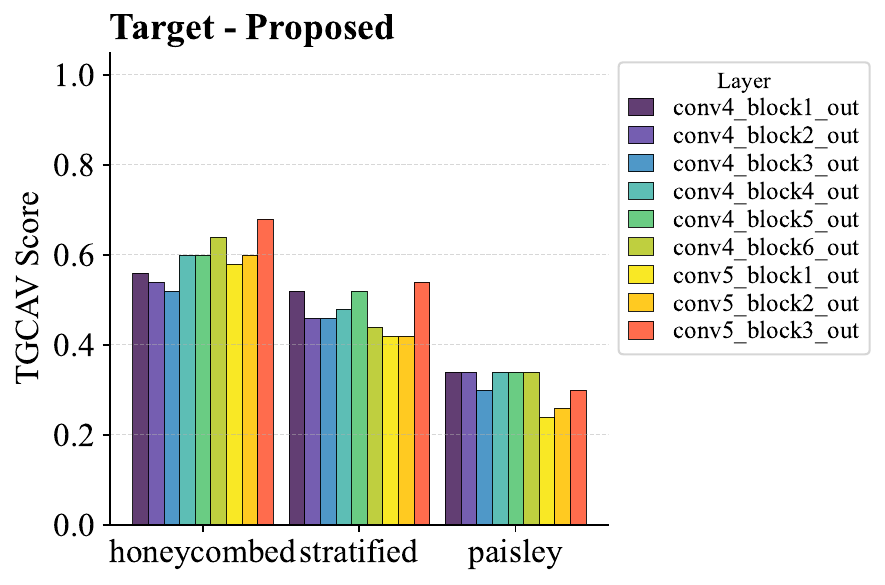}
    \end{subfigure}
    \caption{Bar Chart of TCAV scores of ResNet50V2.}
    \label{fig:tcav bar ResNet50V2}
\end{figure}

Figures~\ref{fig:tcav bar googlenet}, \ref{fig:tcav bar MobileNetV2}, and \ref{fig:tcav bar ResNet50V2} illustrate the distribution of TCAV and TGCAV scores for different concepts across layers. 

\begin{itemize}
    \item \textbf{GoogleNet (Figure~\ref{fig:tcav bar googlenet})}: 
    The top row (original TCAV results) shows that the scores fluctuate significantly across layers, making interpretation inconsistent. For instance, the concept ``cobwebbed'' exhibits high scores in some layers while being nearly absent in others. The bottom row (TGCAV results) demonstrates that our GCAV method stabilizes concept attribution across layers, making it more robust and reliable.

    \item \textbf{MobileNetV2 (Figure~\ref{fig:tcav bar MobileNetV2})}: 
    Similar to GoogleNet, the original TCAV scores exhibit large variations. Notably, the ``striped'' concept has extremely high TCAV scores in some layers but is nearly absent in others. With TGCAV, the concept scores become more evenly distributed, indicating a more stable interpretation.

    \item \textbf{ResNet50V2 (Figure~\ref{fig:tcav bar ResNet50V2})}: 
    The original TCAV scores fluctuate across layers, with some concepts like ``paisley'' exhibiting disproportionately high scores in certain layers. The TGCAV results, however, show that concept influence is distributed more consistently across the network, reducing layer dependence.
\end{itemize}


\subsection{Violin Plot Analysis}
\begin{figure}[h]
    \centering
    \begin{subfigure}{0.33\textwidth}
        \centering
        \includegraphics[width=\textwidth]{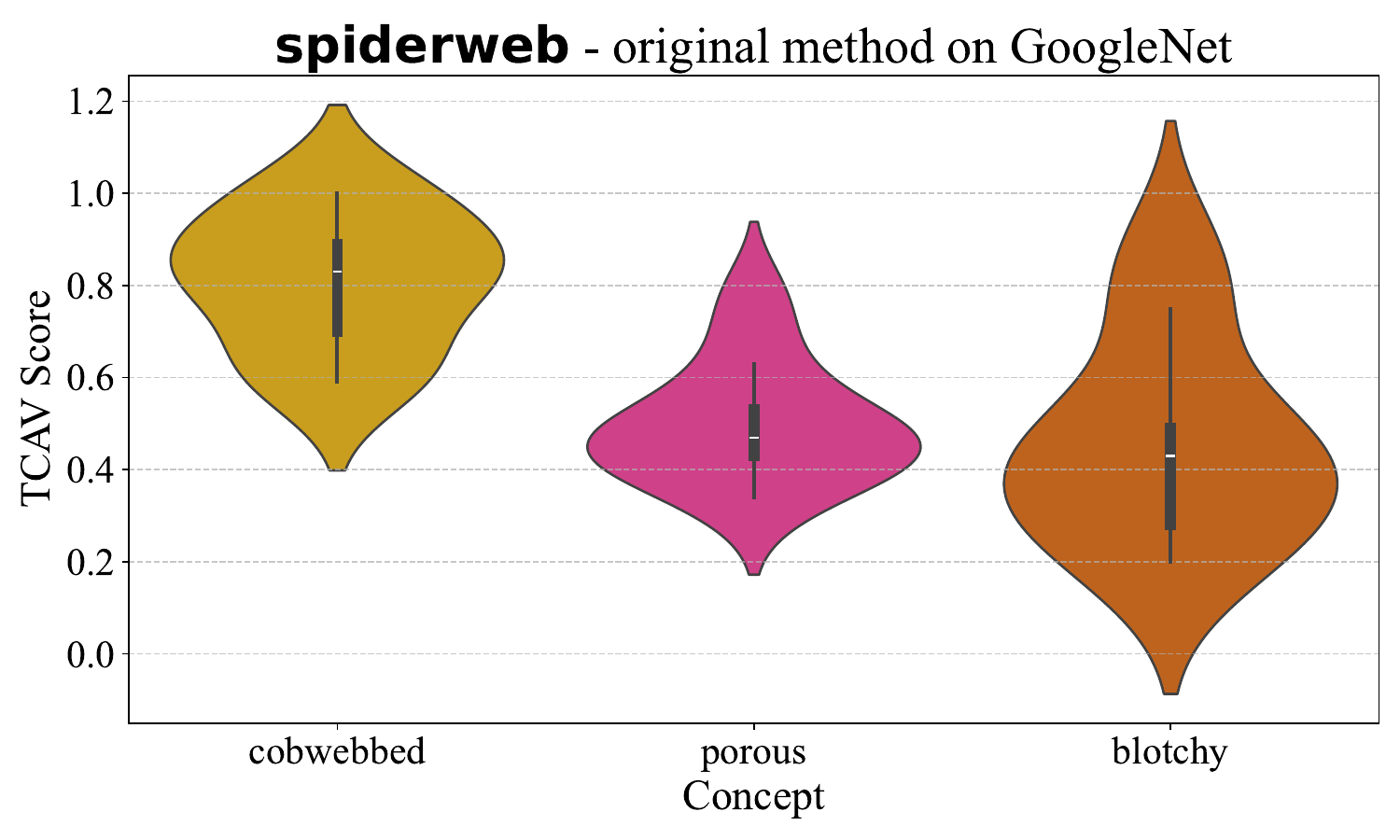}
        \includegraphics[width=\textwidth]{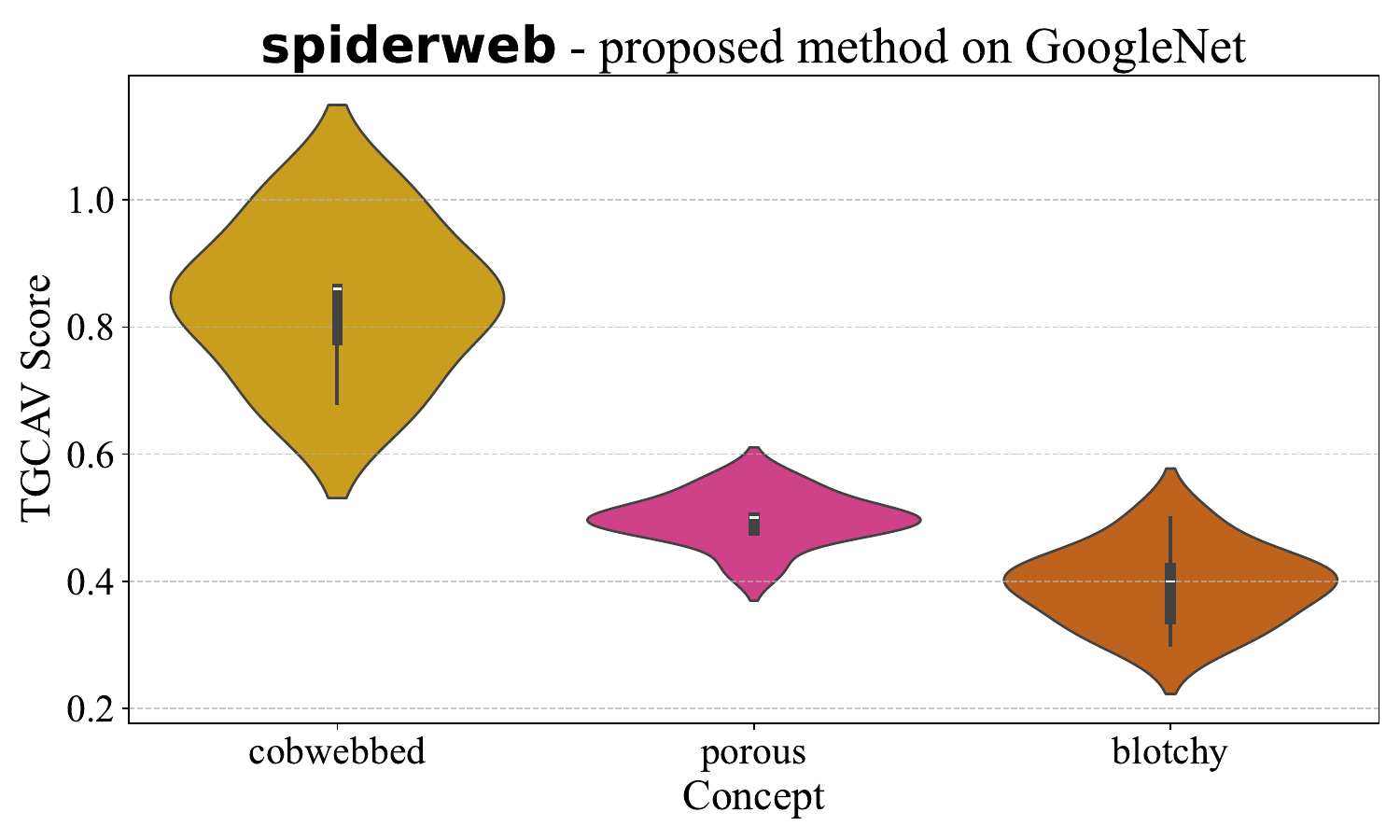}
    \end{subfigure}
    \begin{subfigure}{0.33\textwidth}
        \centering
        \includegraphics[width=\textwidth]{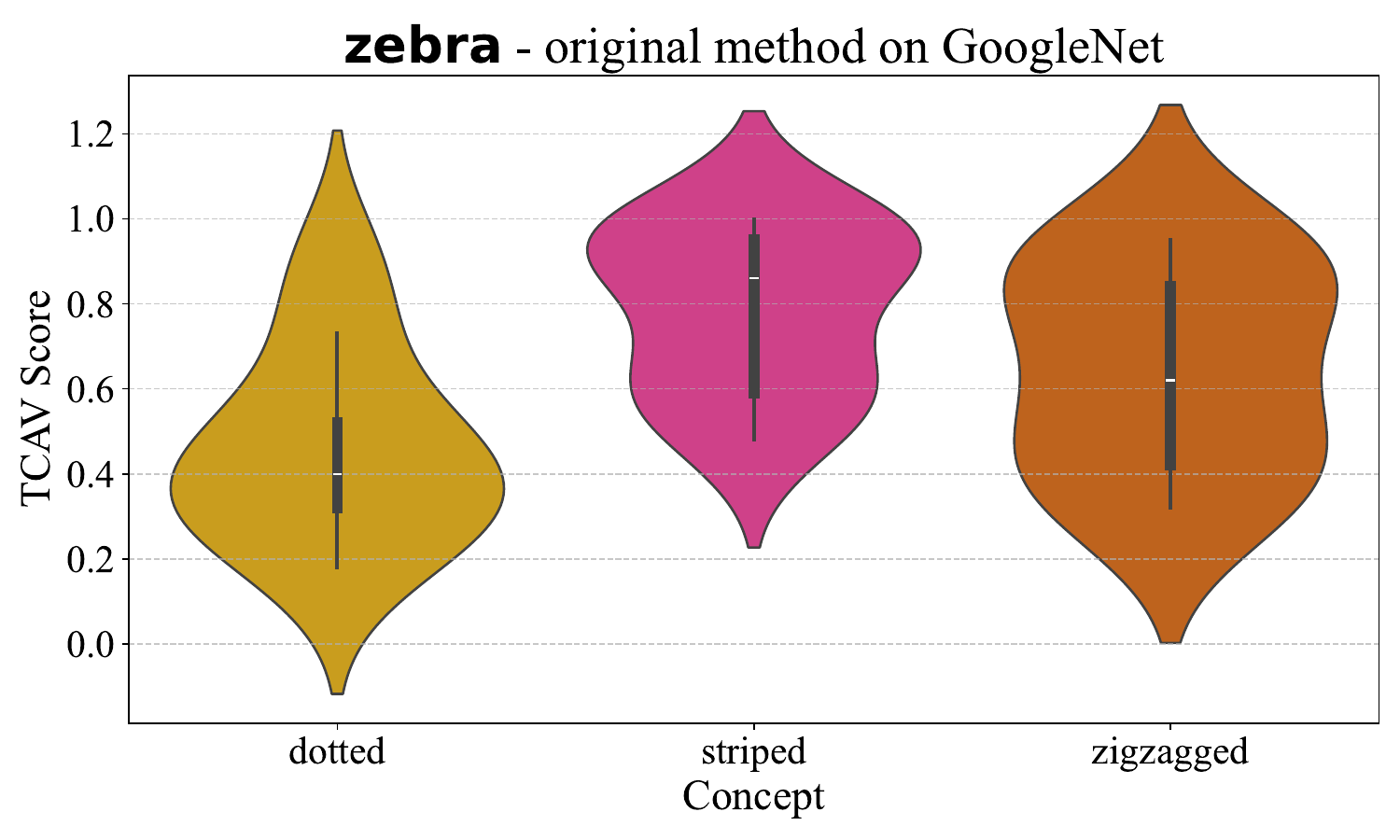}
        \includegraphics[width=\textwidth]{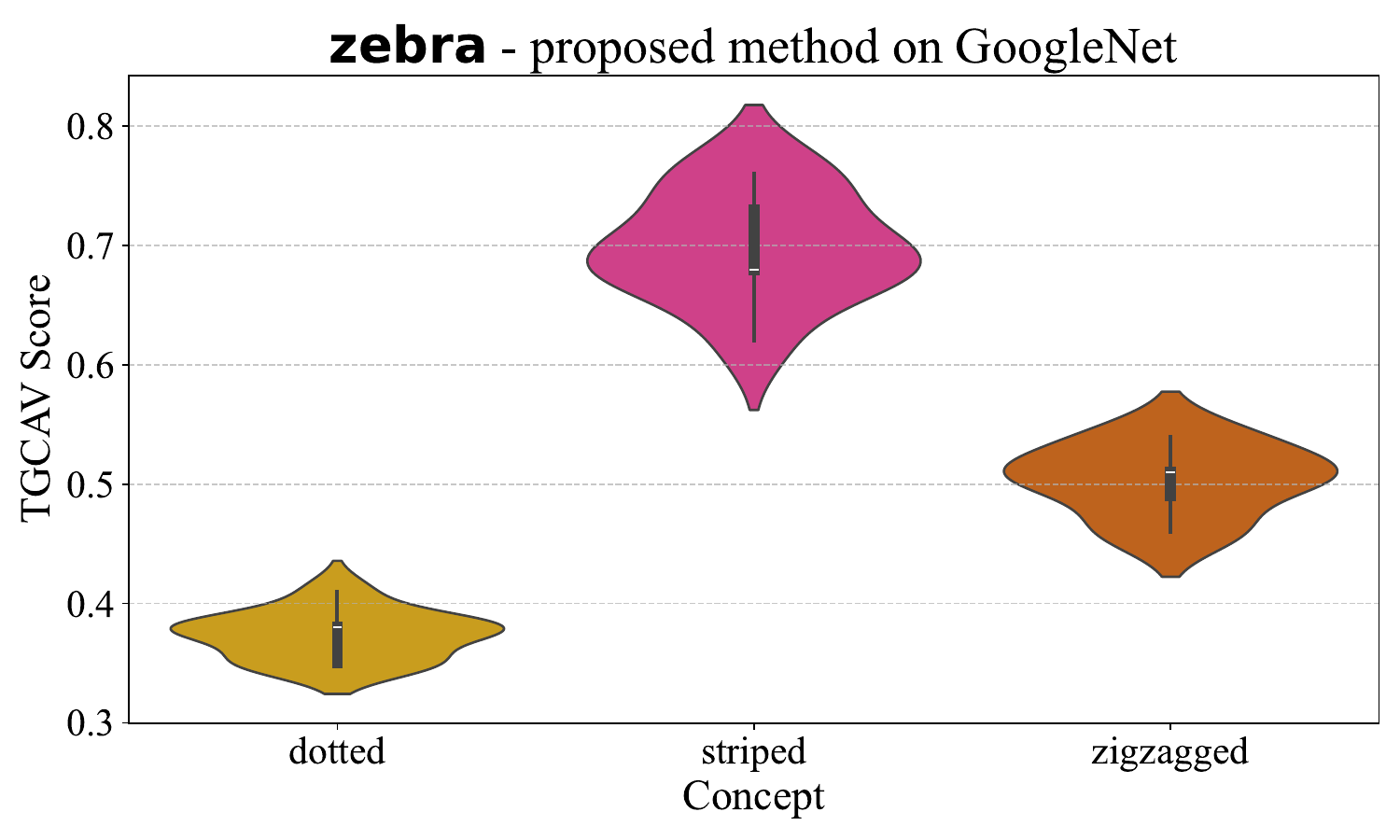}
    \end{subfigure}
        \begin{subfigure}{0.33\textwidth}
        \centering
        \includegraphics[width=\textwidth]{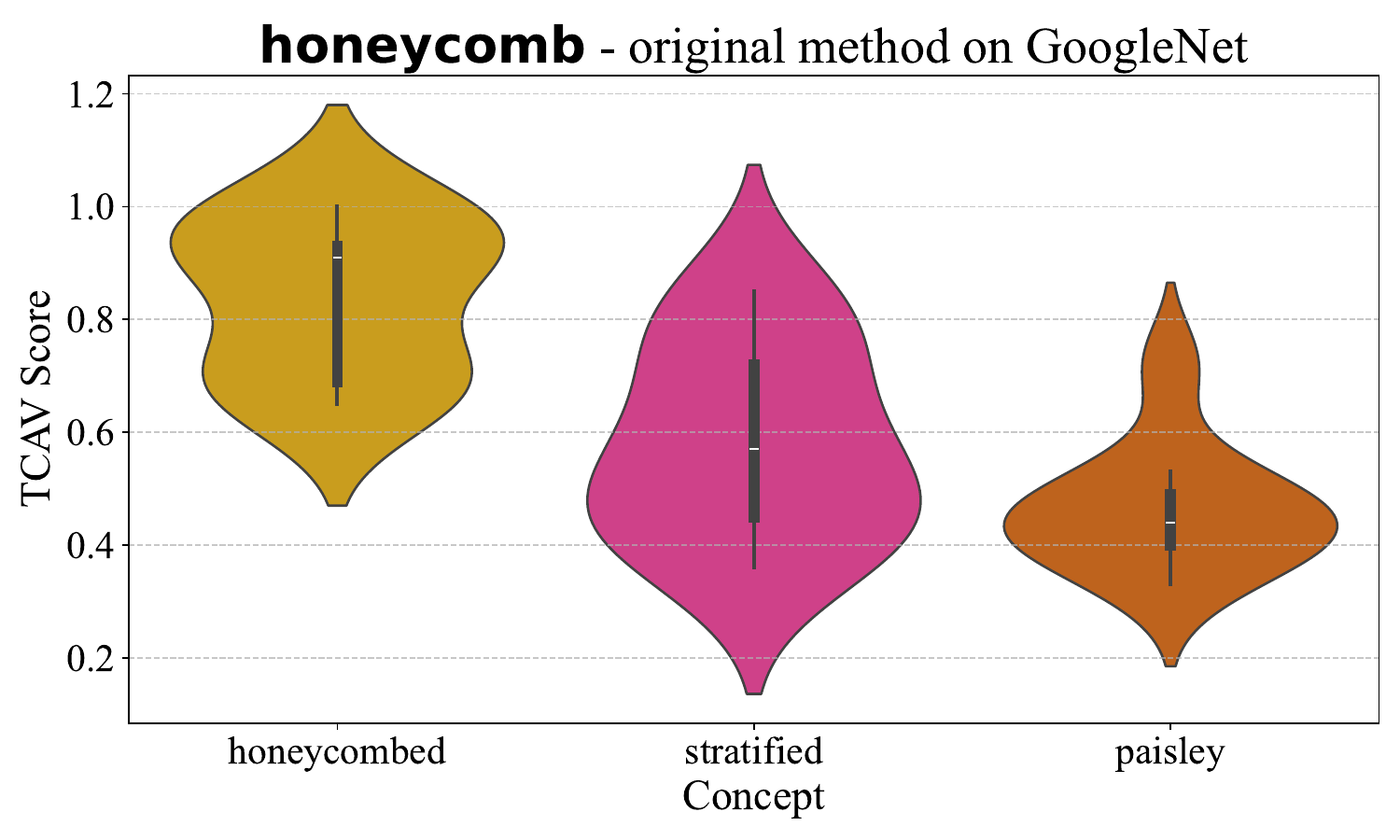}
        \includegraphics[width=\textwidth]{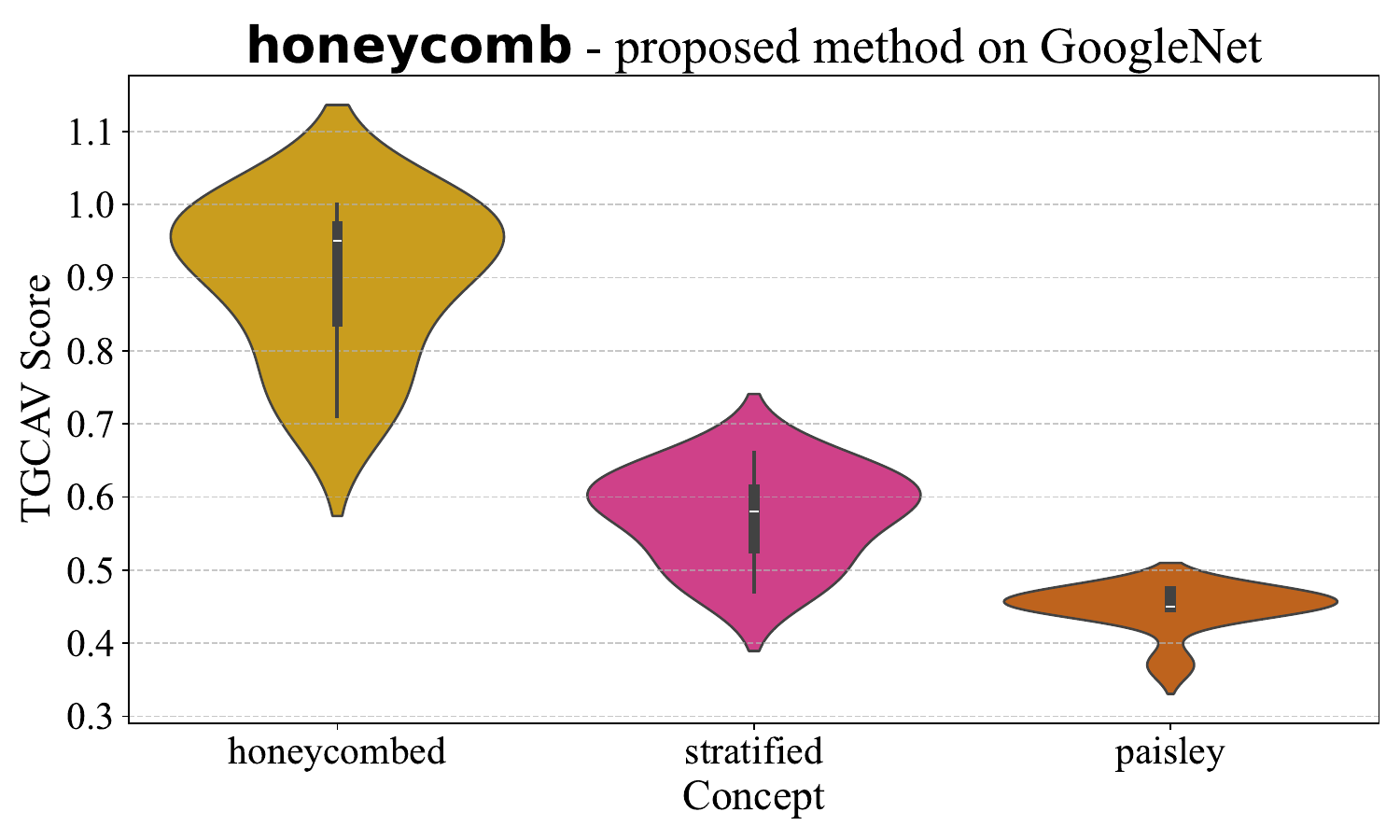}
    \end{subfigure}
    \caption{Violin Plots of TCAV scores of GoogleNet.}
    \label{fig:tcav violin googlenet}
\end{figure}

\begin{figure}[h]
    \centering
    \begin{subfigure}{0.33\textwidth}
        \centering
        \includegraphics[width=\textwidth]{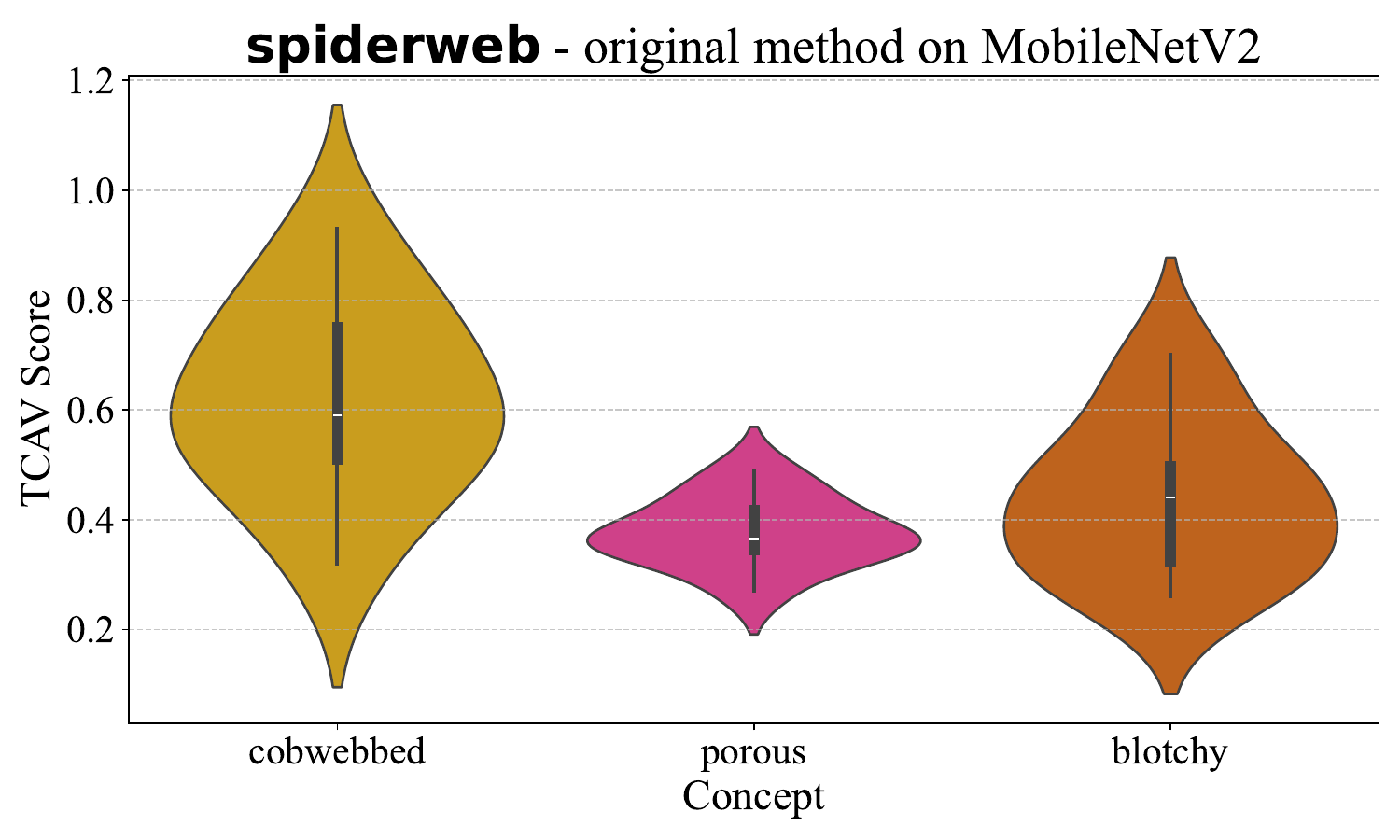}
        \includegraphics[width=\textwidth]{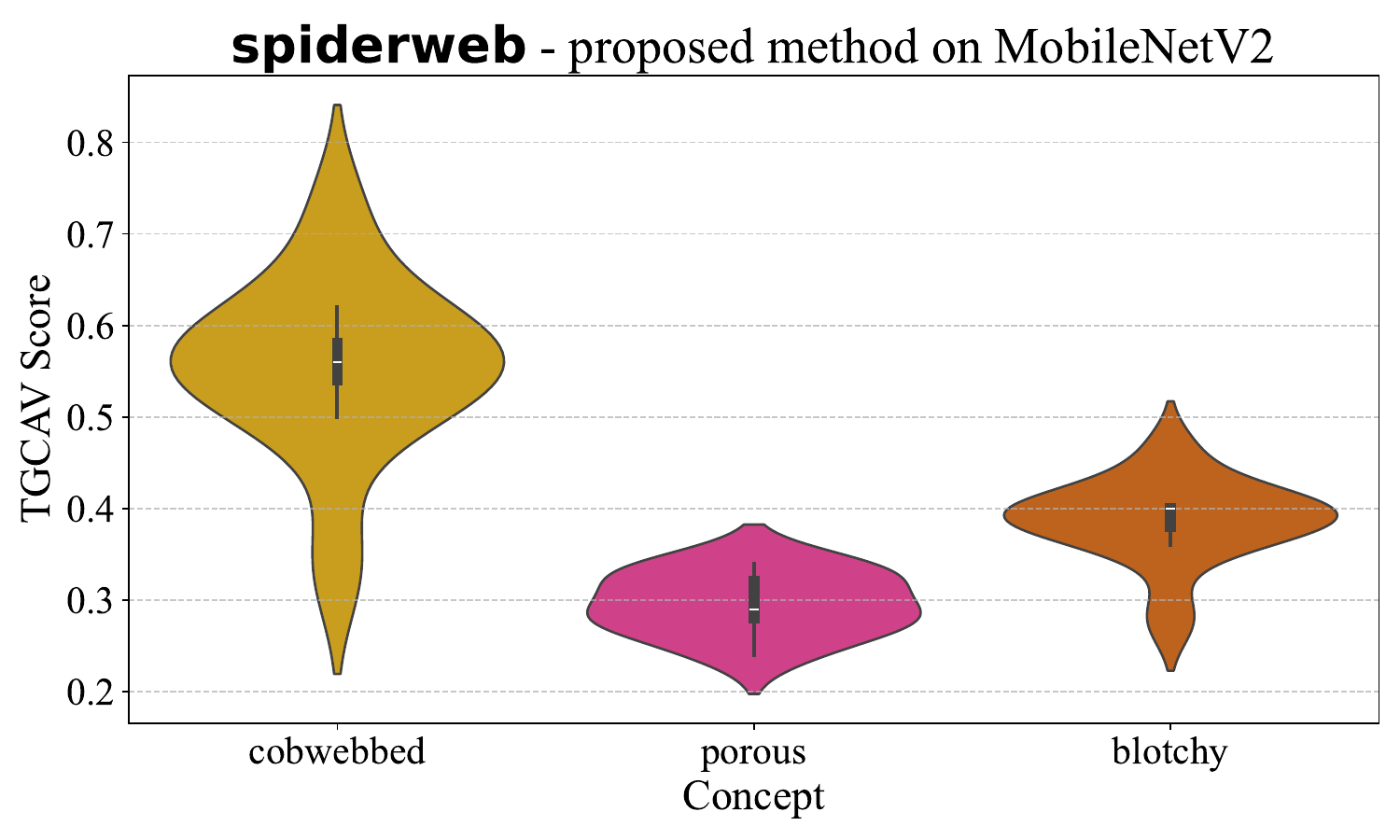}
    \end{subfigure}
    \begin{subfigure}{0.33\textwidth}
        \centering
        \includegraphics[width=\textwidth]{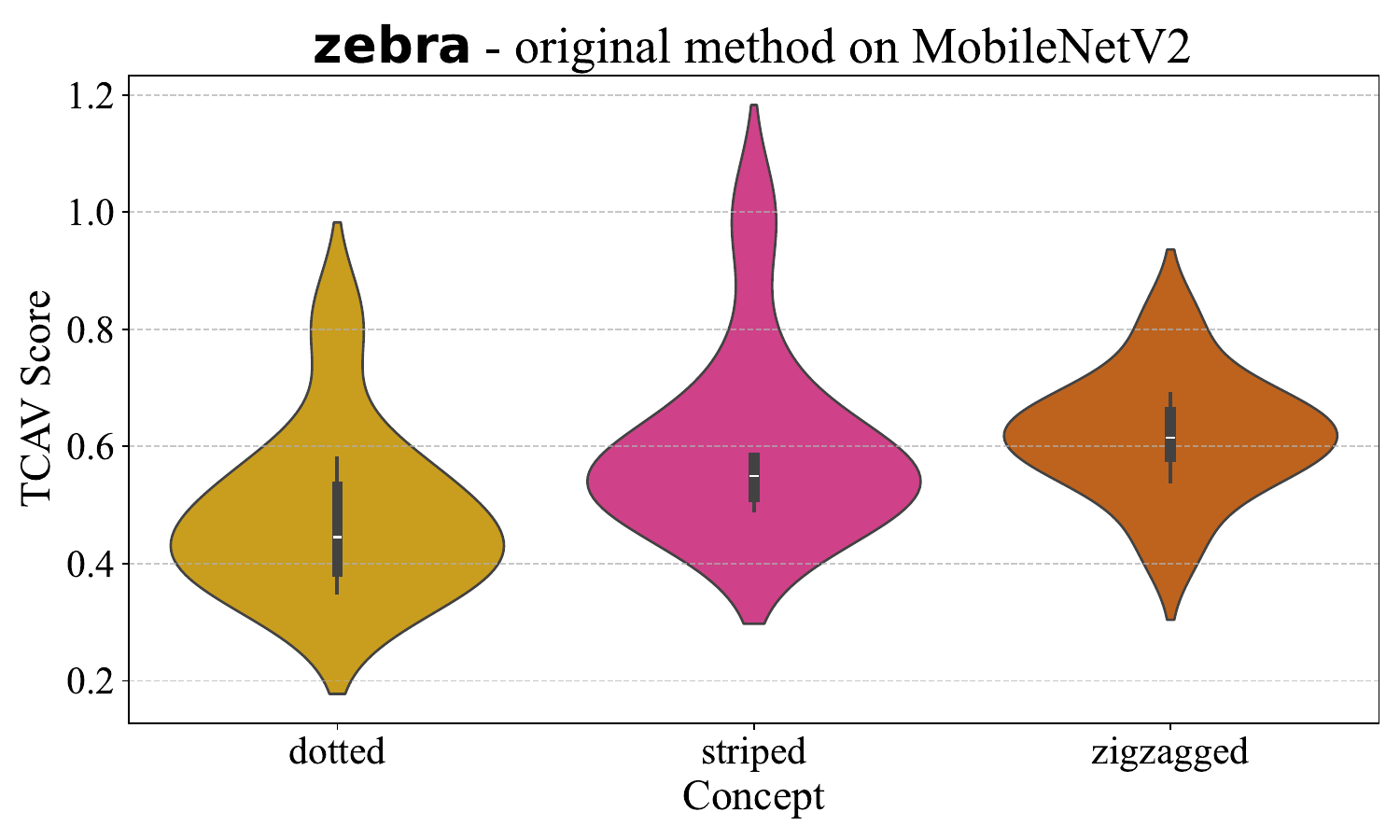}
        \includegraphics[width=\textwidth]{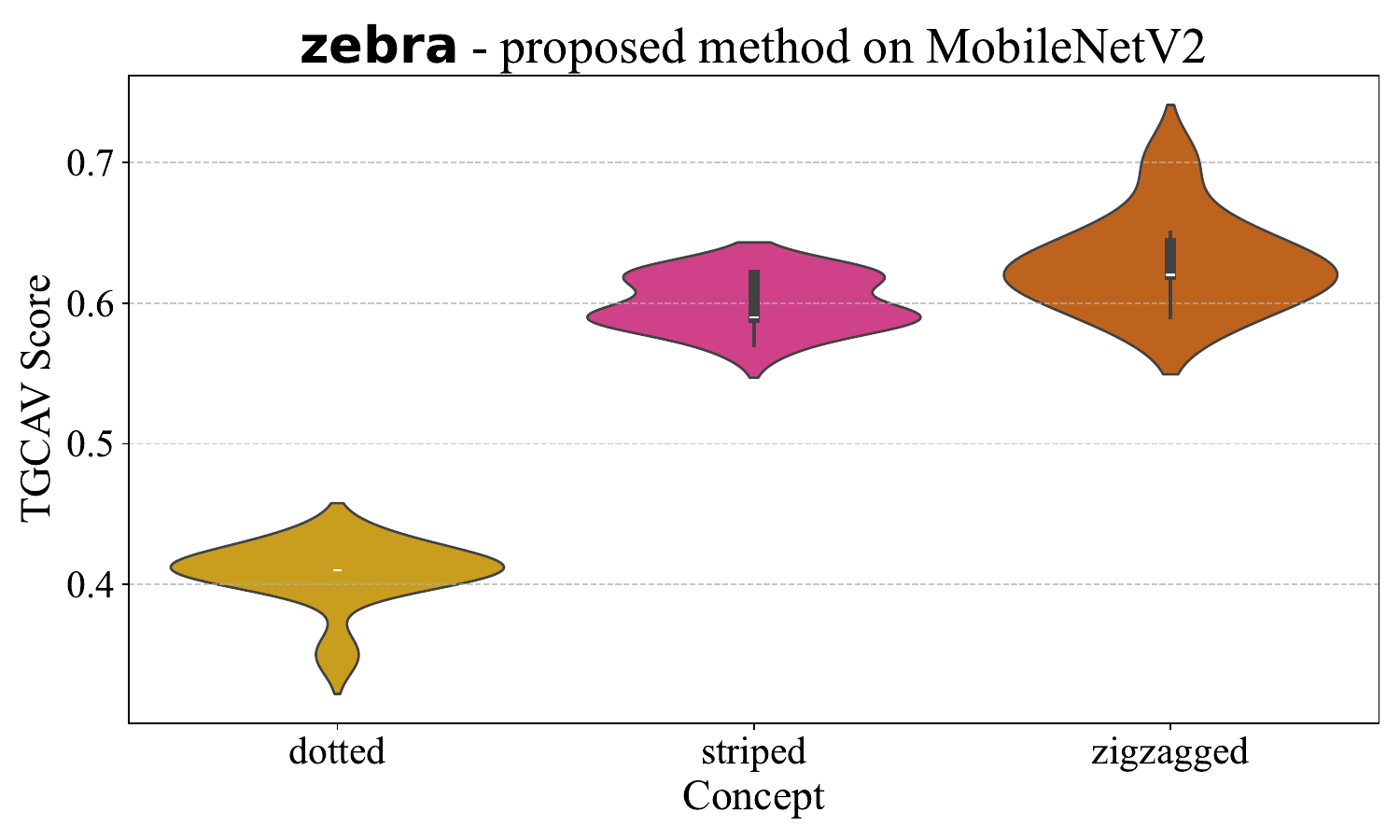}
    \end{subfigure}
        \begin{subfigure}{0.33\textwidth}
        \centering
        \includegraphics[width=\textwidth]{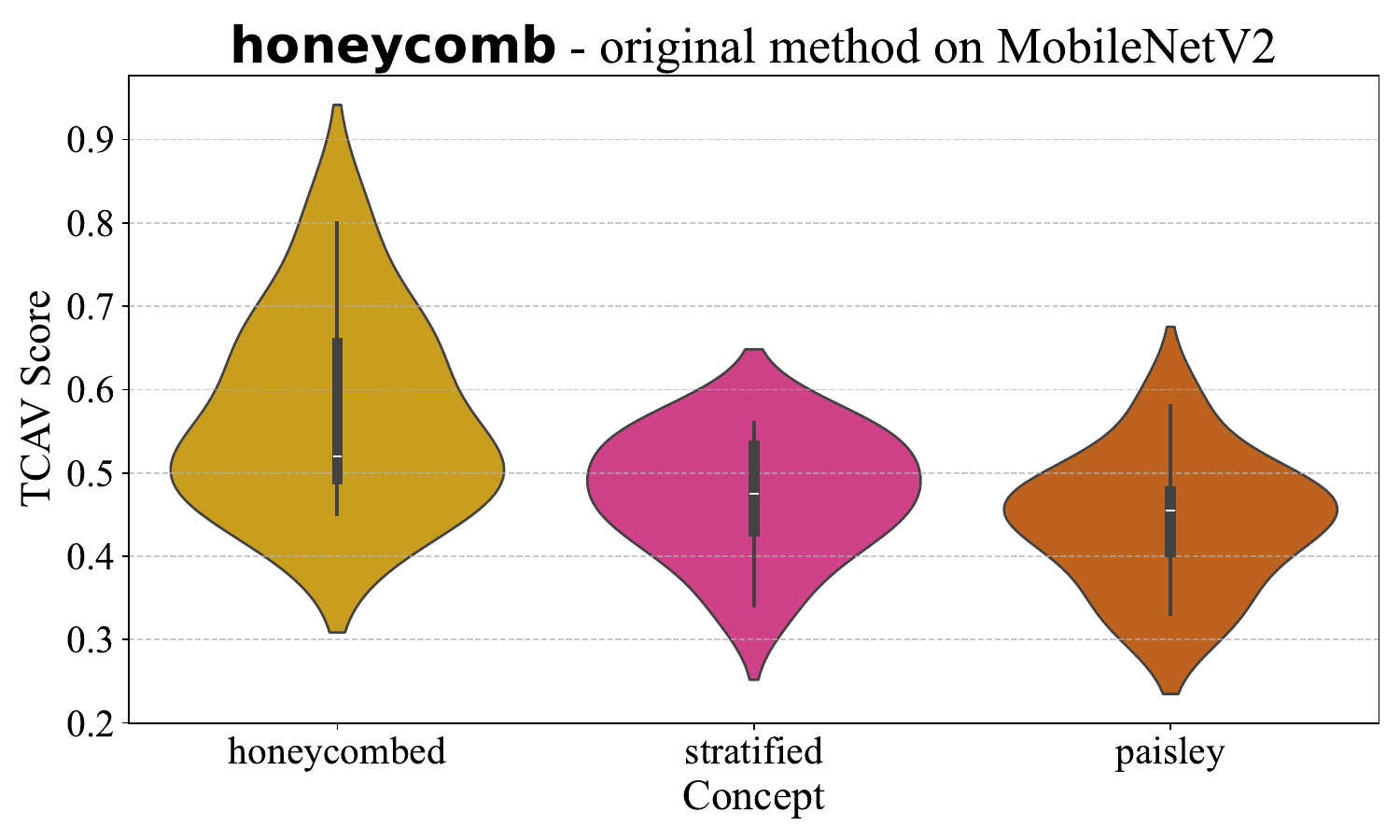}
        \includegraphics[width=\textwidth]{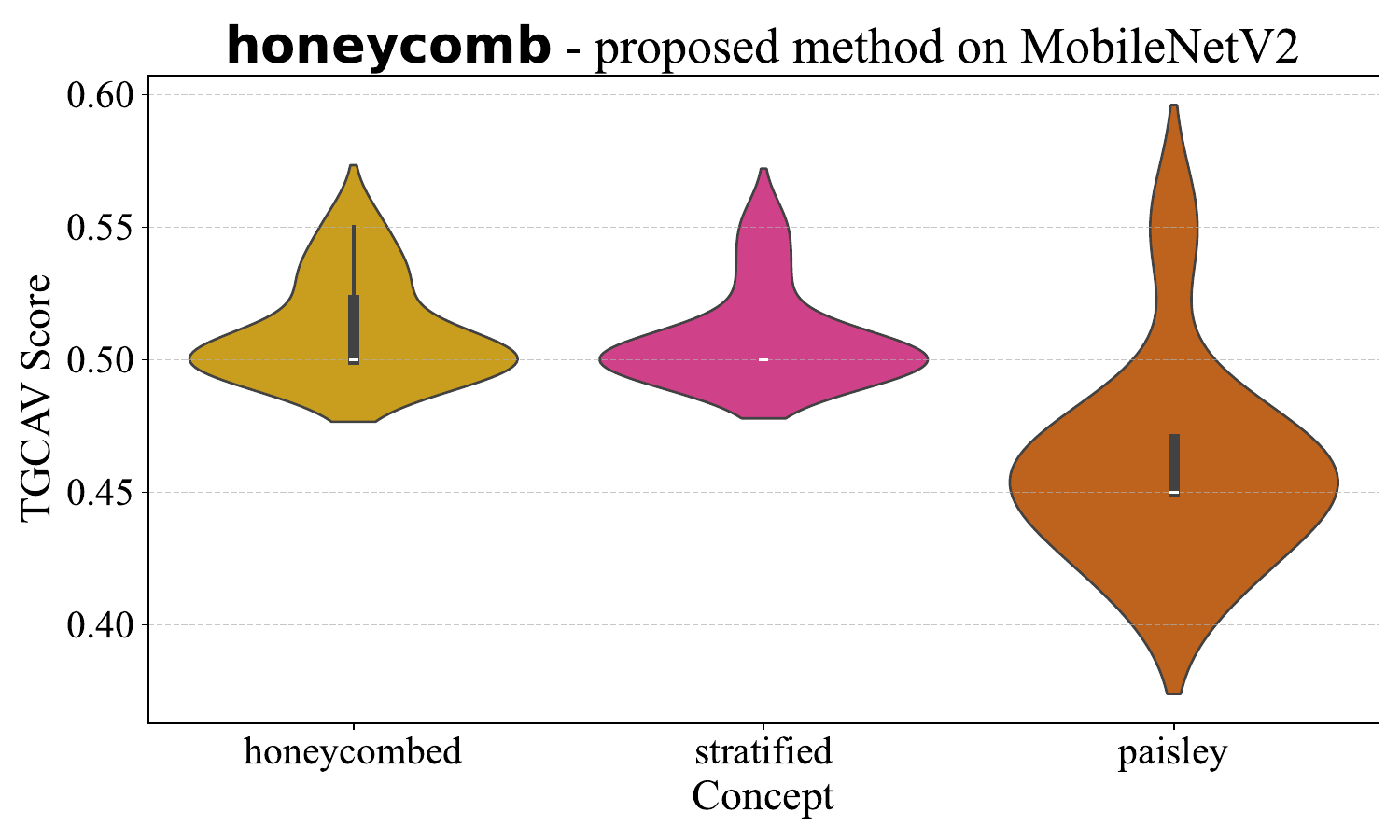}
    \end{subfigure}
    \caption{Violin Plots of TCAV scores of MobileNetV2.}
    \label{fig:tcav violin MobileNetV2}
\end{figure}

\begin{figure}[h]
    \centering
    \begin{subfigure}{0.33\textwidth}
        \centering
        \includegraphics[width=\textwidth]{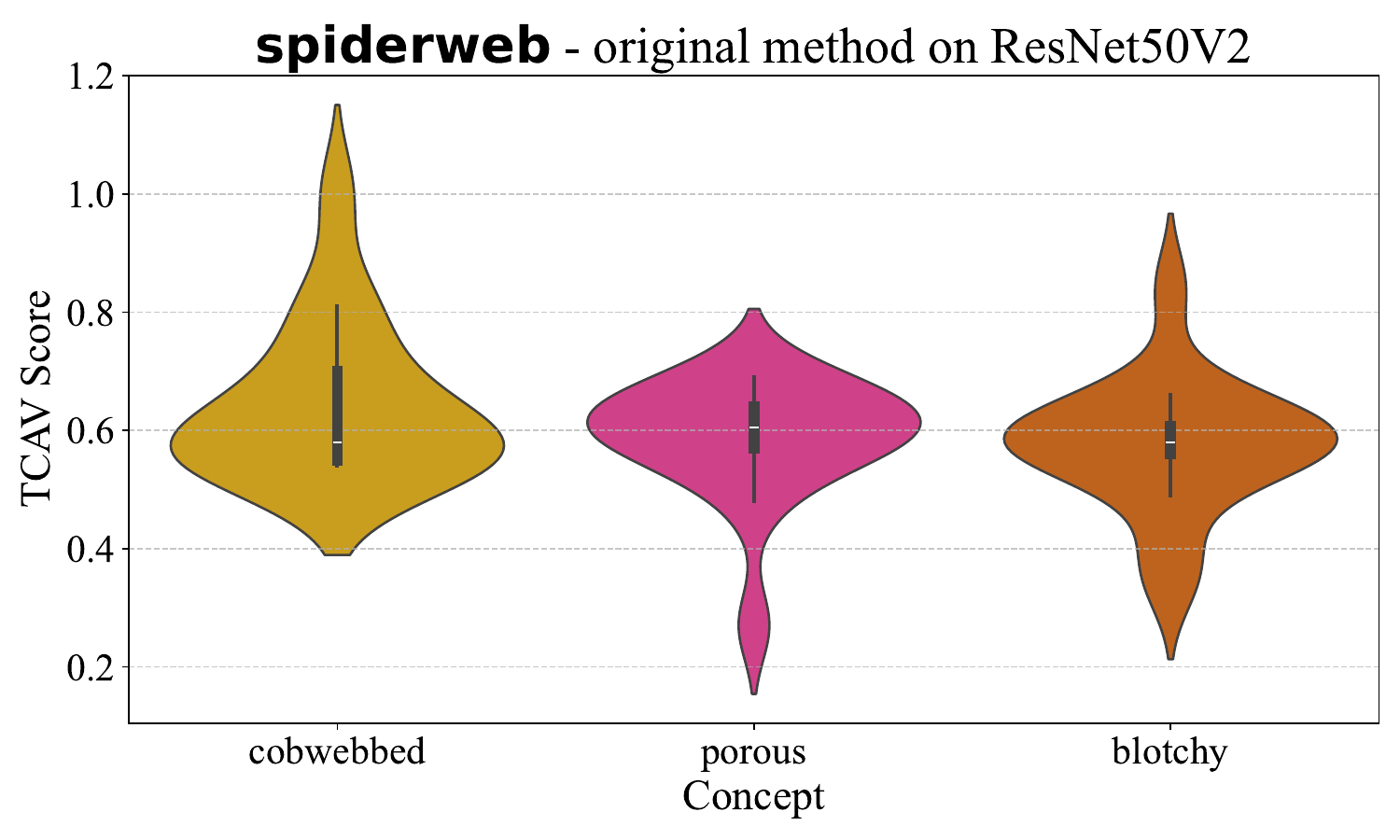}
        \includegraphics[width=\textwidth]{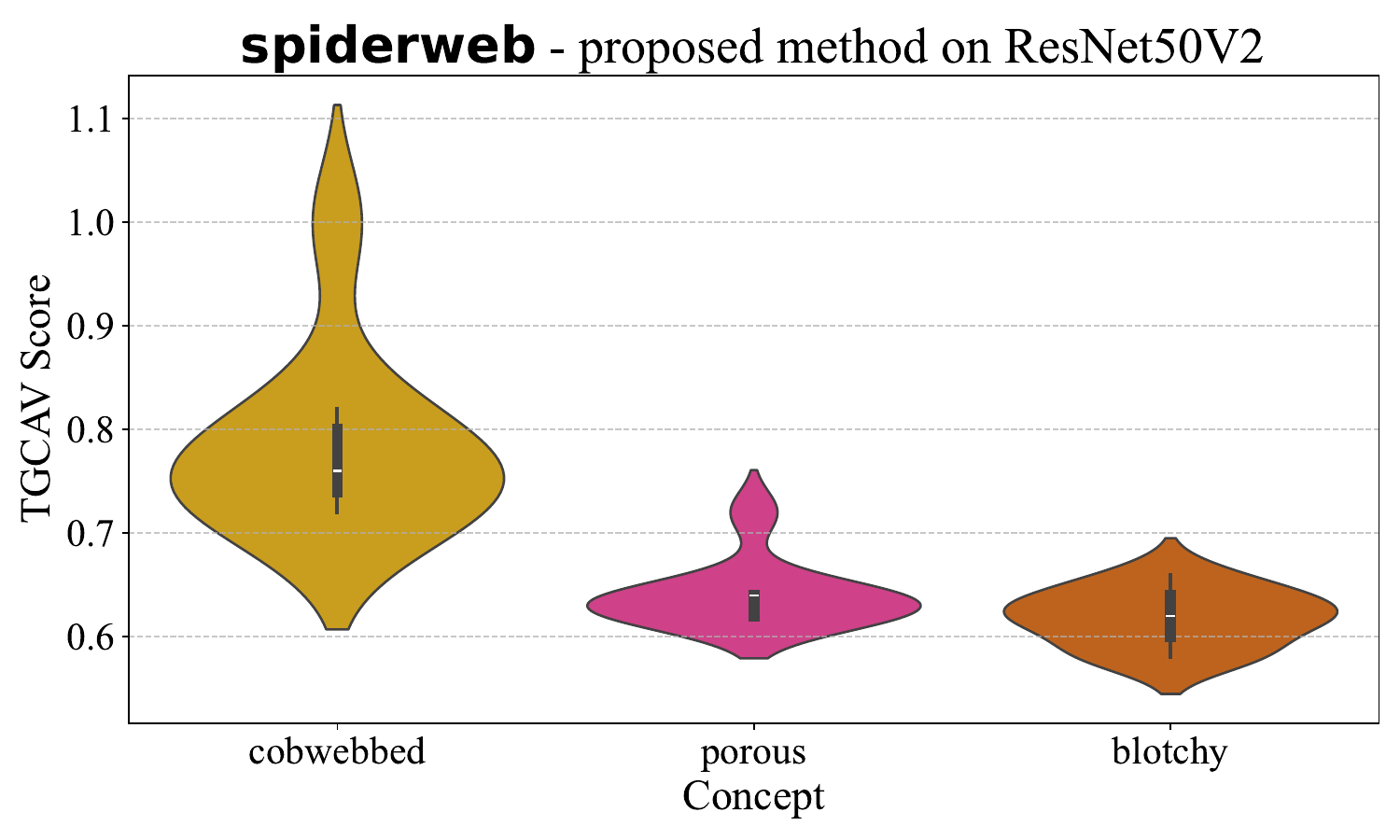}
    \end{subfigure}
    \begin{subfigure}{0.33\textwidth}
        \centering
        \includegraphics[width=\textwidth]{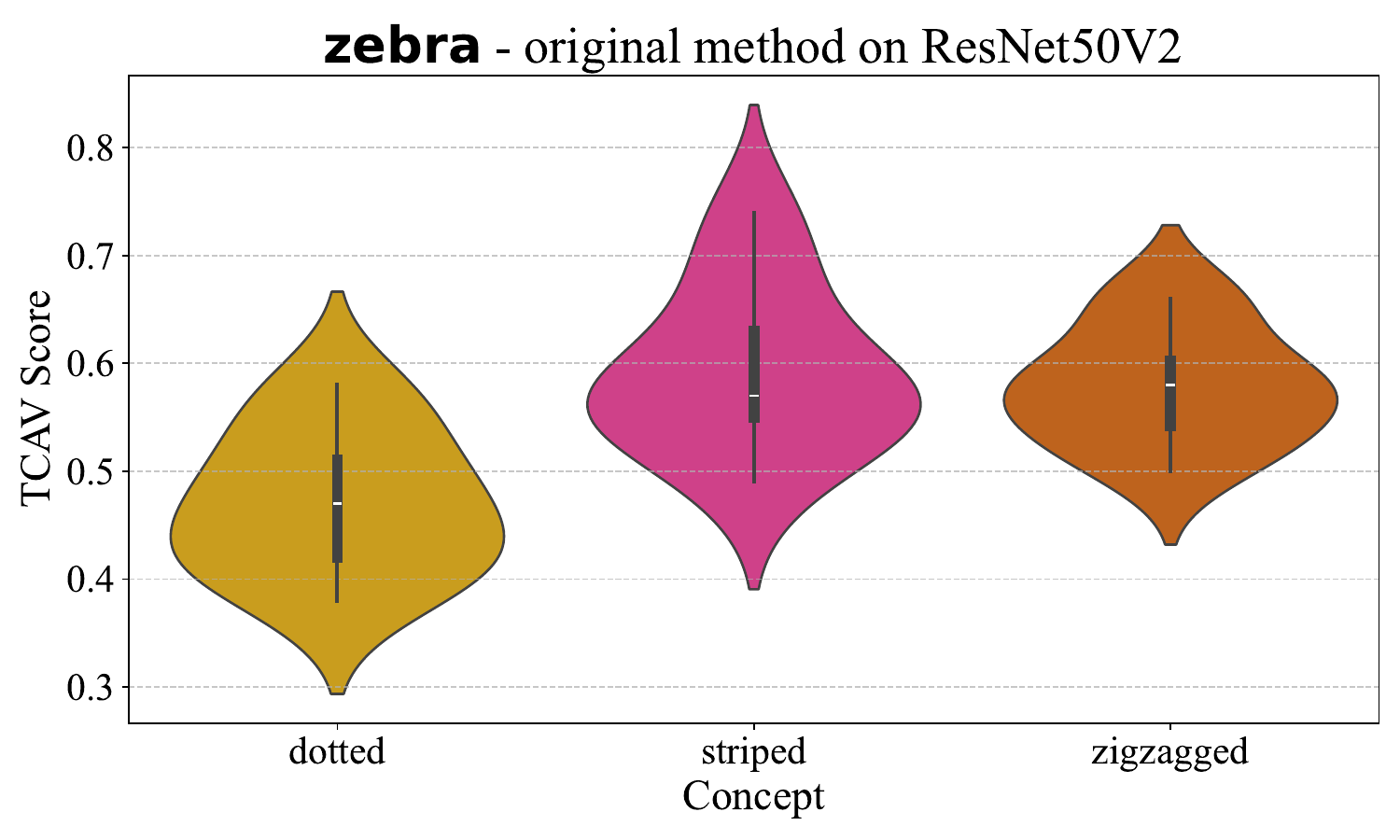}
        \includegraphics[width=\textwidth]{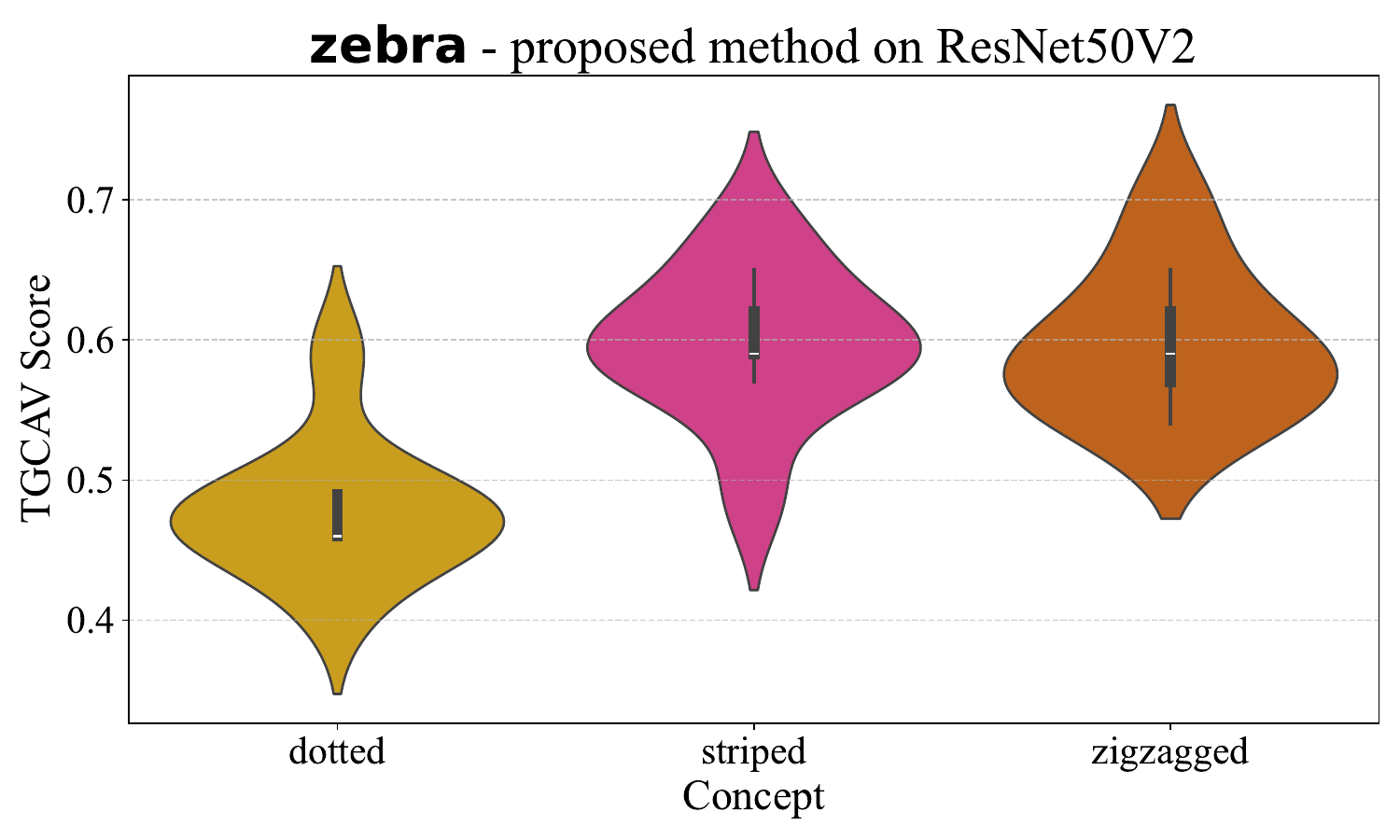}
    \end{subfigure}
        \begin{subfigure}{0.33\textwidth}
        \centering
        \includegraphics[width=\textwidth]{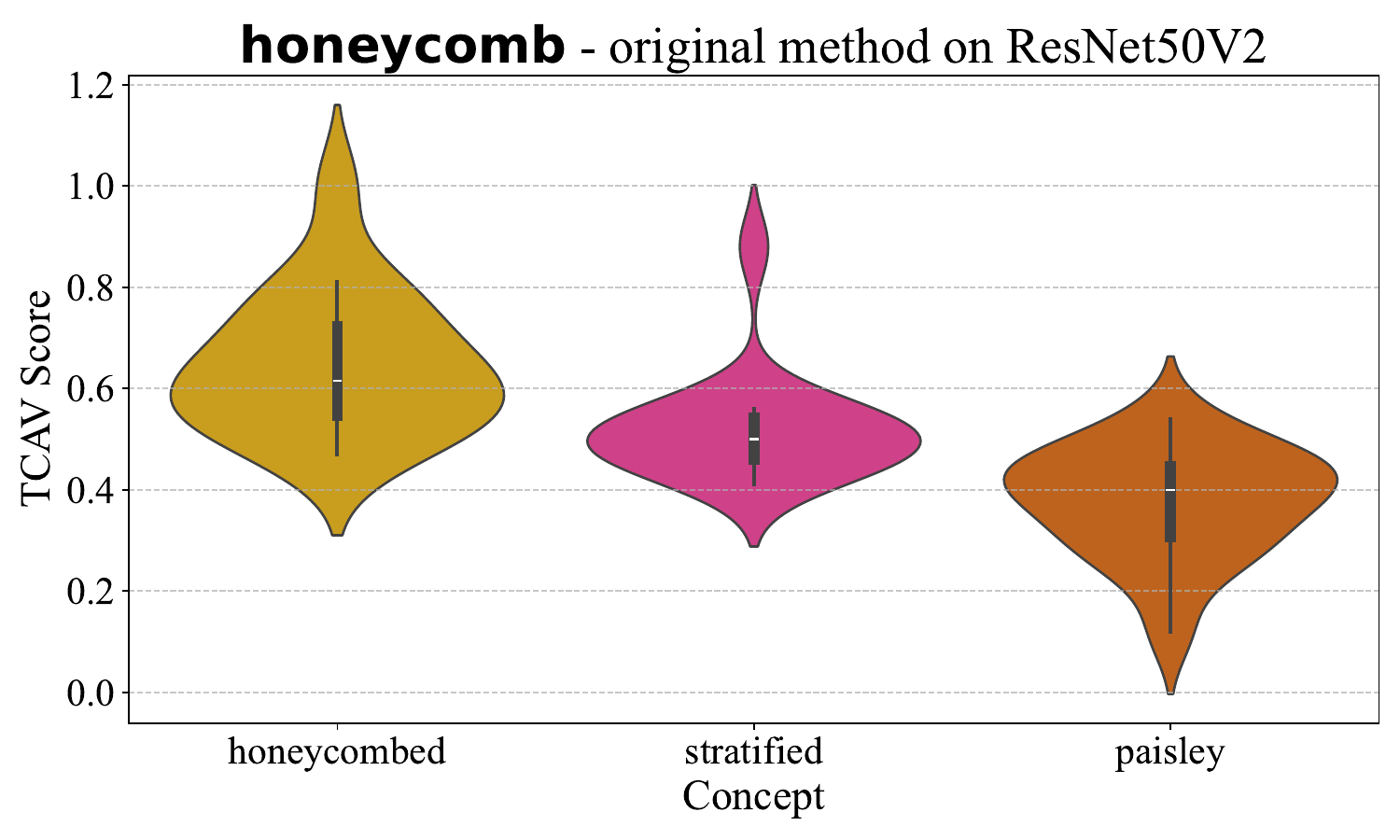}
        \includegraphics[width=\textwidth]{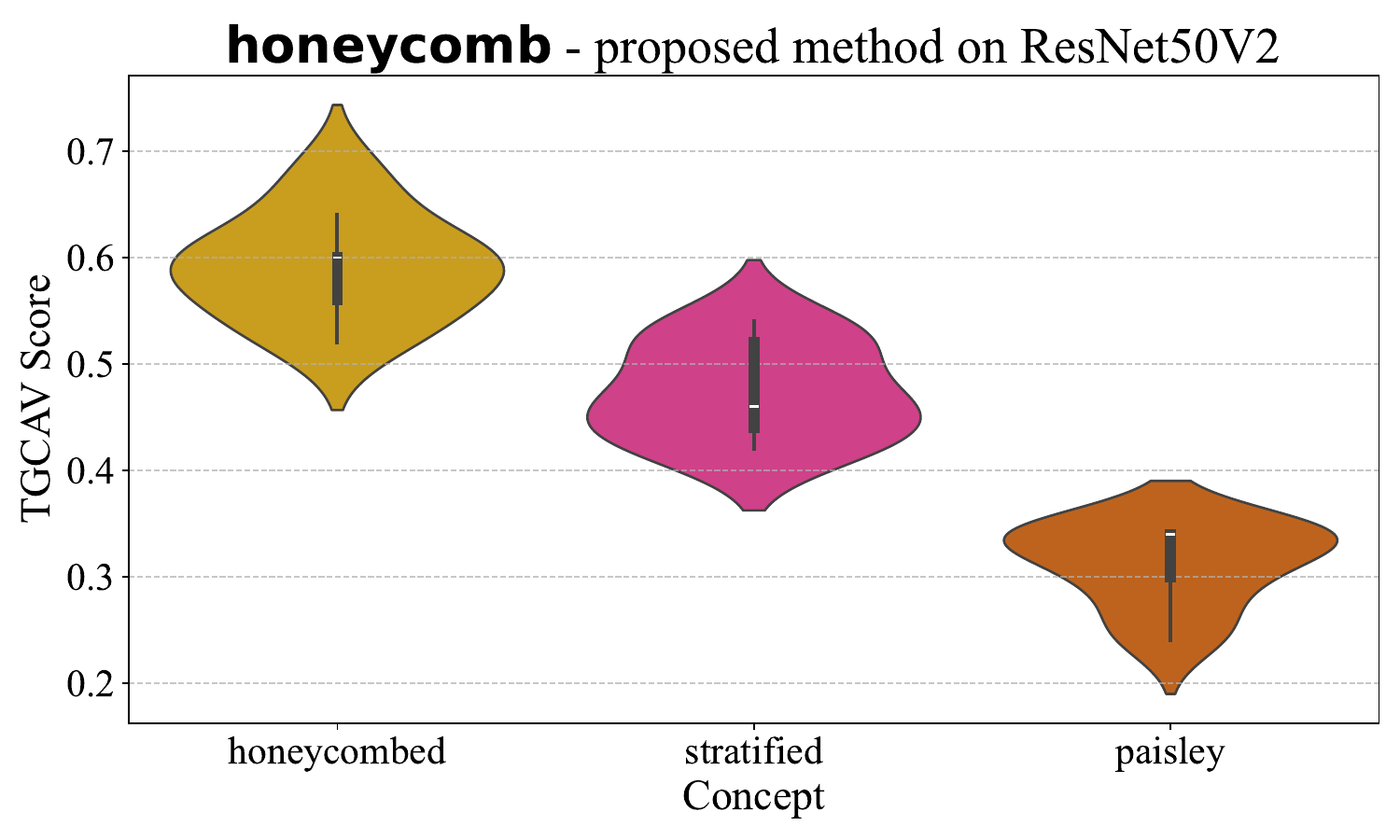}
    \end{subfigure}
    \caption{Violin Plots of TCAV scores of ResNet50V2.}
    \label{fig:tcav violin ResNet50V2}
\end{figure}

Figures~\ref{fig:tcav violin googlenet}, \ref{fig:tcav violin MobileNetV2}, and \ref{fig:tcav violin ResNet50V2} present violin plots that visualize the distribution of TCAV and TGCAV scores across layers. 

\begin{itemize}
    \item \textbf{GoogleNet (Figure~\ref{fig:tcav violin googlenet})}: 
    The original TCAV scores (top row) display high variance, with some concepts exhibiting a bimodal distribution, indicating instability across layers. For example, the ``zigzagged'' concept has strong activations in certain layers while being much weaker in others. The TGCAV results (bottom row) show a much more compact distribution, confirming the stabilization effect of our GCAV framework.

    \item \textbf{MobileNetV2 (Figure~\ref{fig:tcav violin MobileNetV2})}: 
    The original TCAV scores exhibit large fluctuations across layers, with some concepts like ``honeycombed'' showing extreme variations. After applying GCAV, the distributions become significantly more compact, ensuring that concept influence is more evenly spread across the layers.

    \item \textbf{ResNet50V2 (Figure~\ref{fig:tcav violin ResNet50V2})}: 
    The original TCAV distributions display significant variability, highlighting the instability of layer-wise concept attribution. The TGCAV results demonstrate that our method successfully reduces variance, making concept attributions more stable across layers.
\end{itemize}

\subsection{Summary}
\begin{itemize}
    \item The \textbf{bar charts} show that TGCAV significantly reduces the cross-layer variability of TCAV scores, ensuring more stable concept interpretations.
    \item The \textbf{violin plots} further illustrate that TGCAV leads to more compact distributions, reducing variance and enhancing consistency in concept activations.
    \item Overall, GCAV improves the reliability of concept-based interpretability by integrating cross-layer information, reducing layer selection uncertainty, and enhancing robustness against adversarial perturbations.
\end{itemize}

\end{document}